\documentclass[10pt,twocolumn,letterpaper]{article}

\usepackage{cvpr}              
\usepackage[table]{xcolor}
\usepackage{multirow}
\usepackage{rotating}
\usepackage{fontawesome}
\definecolor{cvprblue}{rgb}{0.21,0.49,0.74}
\usepackage[pagebackref,breaklinks,colorlinks,allcolors=cvprblue]{hyperref}

\def\paperID{18484} 
\def\confName{CVPR}
\def\confYear{2025}

\title{Less is More: Efficient Model Merging with Binary Task Switch}
\vspace{-10pt}
\author{
  Biqing Qi\textsuperscript{1}, 
  Fangyuan Li\textsuperscript{2},
  Zhen Wang\textsuperscript{3},
  Junqi Gao\textsuperscript{1,3,}\thanks{Corresponding authors.},
  Dong Li\textsuperscript{1,3},
  Peng Ye\textsuperscript{1}, 
  Bowen Zhou\textsuperscript{1,4,}\footnotemark[1] \\
  $^1$ Shanghai Artificial Intelligence Laboratory, \\
  $^2$ Department of Control Science and Engineering, Harbin Institute of Technology, \\
  $^3$ School of Mathematics, Harbin Institute of Technology, \\
  $^4$ Department of Electronic Engineering, Tsinghua University \\
  {\tt\small \{qibiqing7,jacklee19900212,wz443534070,gjunqi97,arvinlee826\}@gmail.com,} \\ 
  {\tt\small 20110720039@fudan.edu.cn,zhoubowen@tsinghua.edu.cn}
  }

\begin{document}
\maketitle
\vspace{-10pt}
\begin{abstract}
\vspace{-5pt}

As an effective approach to equip models with multi-task capabilities without additional training, model merging has garnered significant attention. However, existing methods face challenges of redundant parameter conflicts and the excessive storage burden of parameters. In this work, through controlled experiments, we reveal that for task vectors, only those parameters with magnitudes above a certain threshold contribute positively to the task, exhibiting a pulse-like characteristic. We then attempt leveraging this characteristic to binarize the task vectors and reduce storage overhead. Further controlled experiments show that the binarized task vectors incur almost no decrease in fine-tuning and merging performance, and even exhibit stronger performance improvements as the proportion of redundant parameters increases. Based on these insights, we propose Task Switch (T-Switch), which decomposes task vectors into three components: 1) an activation switch instantiated by a binarized mask vector, 2) a polarity switch instantiated by a binarized sign vector, and 3) a scaling knob instantiated by a scalar coefficient. By storing task vectors in a binarized form, T-Switch alleviates parameter conflicts while ensuring efficient task parameter storage. Furthermore, to enable automated switch combination in T-Switch, we further introduce Auto-Switch, which enables training-free switch combination via retrieval from a small query set. Experiments indicate that our methods achieve significant performance improvements over existing baselines, requiring only 1-3$\%$ of the storage space of full-precision parameters.

\end{abstract}
\vspace{-5pt}

\section{Introduction}
\label{sec:intro}
With the thriving of open-source communities \cite{abs-1910-03771,chen2019mmdetection}, a growing number of pre-trained and fine-tuned models are being widely used \cite{touvron2023llama,bai2023qwen}. Directly leveraging these models to address specific tasks becomes a mainstream practice. However, facing a multitude of task scenarios, deploying dedicated fine-tuned models for each task incurs significant storage and computational costs, becoming impractical in scenarios with limited resources. To address the challenge of efficiently managing and applying models in multi-task scenarios, model merging \cite{DBLP:conf/iclr/IlharcoRWSHF23,DBLP:conf/nips/MatenaR22} offers a promising solution. By merging parameters of multiple task-specific model, model merging effectively integrates knowledge from different tasks, enhancing the model's adaptability in multi-task settings without additional training.

\begin{figure} [t]
	\centering
\includegraphics[width=0.46\textwidth]{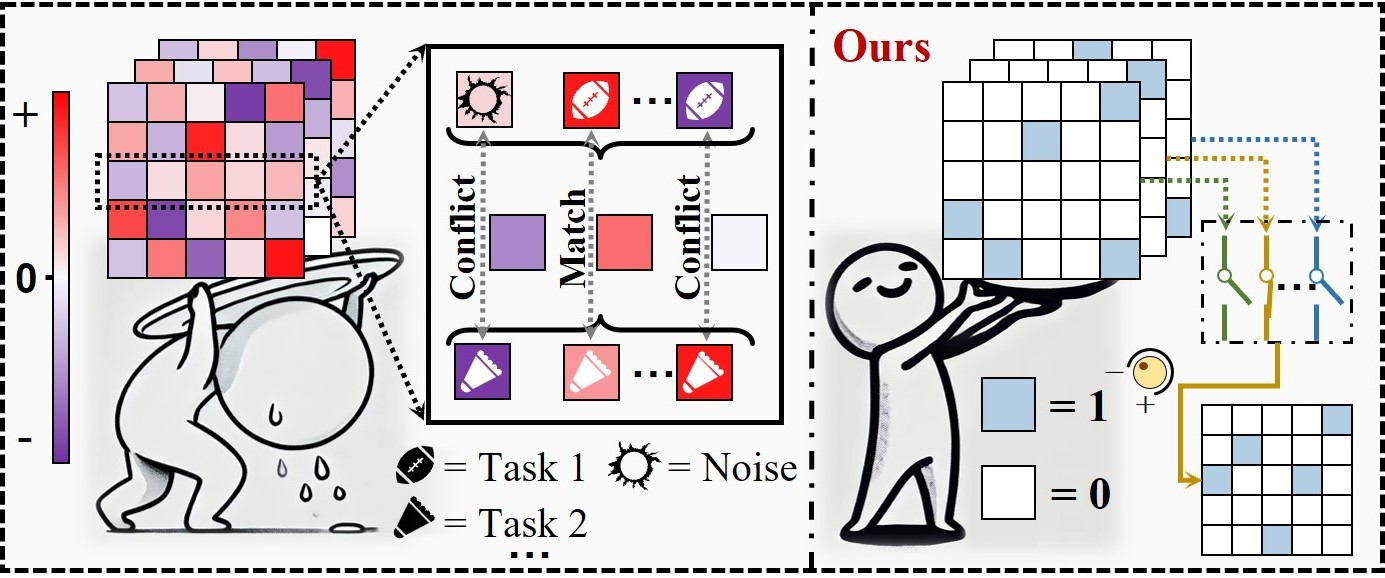}
\vspace{-3pt}
	\caption{Left: Challenges of model merging: conflicts in task vectors and the burden of parameter storage. Right: Our method eliminates redundancy while enabling the storage of binarized, lightweight task vectors.}
	\label{confilct} 
\vspace{-18pt}
\end{figure}

Current model merging approaches can be broadly classified into two types: static merging \cite{DBLP:conf/nips/MatenaR22,DBLP:conf/iclr/Jin0P023,DBLP:conf/iclr/IlharcoRWSHF23} and dynamic merging \cite{DBLP:journals/corr/abs-2406-15479,DBLP:journals/corr/abs-2405-17461}. Static merging strategies, such as Task-Arithmetic \cite{DBLP:conf/iclr/IlharcoRWSHF23}, involve linearly combining the differences between the weights of different fine-tuned models and the pre-trained weights (referred to as task vectors). Techniques like RegMean \cite{DBLP:conf/iclr/Jin0P023} minimize the difference between the parameter matrix and input vector products before and after merging. These methods maintain static weights after merging, but conflicts between task vectors limit their performance, making static merging inadequate for dynamically changing task scenarios. In contrast, dynamic merging methods update the merging strategy flexibly based on task variations. For example, Twin-Merging \cite{DBLP:journals/corr/abs-2406-15479} learns a router from a set of instance data to automatically merge task vectors, while EMR-Merging \cite{DBLP:journals/corr/abs-2405-17461} uses task-specific mask matrices and scaling factors to trim the unified task vector for each task, offering better adaptability to diverse task environments.

Although dynamic merging approaches offer advantages due to their stronger task adaptability, existing strategies still face challenges illustrated in Fig.\ref{confilct}: 1) Conflicts between task vectors limit performance potential. In strategies that first merge task vectors into a unified task vector before applying dynamic pruning, the merging of task vectors is still constrained by their inherent conflicts \cite{DBLP:conf/nips/YadavTCRB23}, which limits the performance gains from further pruning. Even automatic combination methods are also affected by this issue. 2) Storing task vectors imposes significant storage overhead. If all task vectors are stored and automatically combined via a router, the need to store each task vector in full precision \cite{DBLP:journals/corr/abs-2406-15479}, with a parameter count close to that of the original model, results in storage requirements several times that of the pre-trained weights, hindering the application of such methods in resource-constrained scenarios.

Therefore, finding a better balance between performance and storage efficiency is key to the broader adoption of dynamic merging methods in practical applications. In this context, our research aims to explore a dynamic merging solution that can alleviate task vector conflicts while maintaining high storage efficiency.

To alleviate task vector conflicts, we re-examine the issue of parameter redundancy within task vectors. Previous work \cite{DBLP:conf/icml/Yu0Y0L24} shows that a large amount of redundant parameters exist in task vectors, and these parameters may interfere with those that make significant contributions to other tasks during merging. In this work, we investigate the relationship between this redundancy and the parameter magnitudes to provide a methodology for discarding redundant parameters. Specifically, we first design a pulse activation mechanism to control the magnitude and proportion of discarded parameters. Based on this mechanism, we conduct a series of controlled experiments to explore the impact of discarding task vector parameters with different magnitudes on task performance. The experimental results show that parameters only make a significant contribution to the task when their magnitude exceeds a certain activation threshold, and discarding the remaining parameters not only does not affect task accuracy, but can even lead to further performance improvements, exceeding the performance of the original fine-tuned model. This suggests that parameters in the task vector exhibit a pulse-like characteristic, where those with smaller magnitudes are redundant and may negatively impact task performance.

To further improve storage efficiency while alleviating parameter conflicts, we leverage the pulse-like characteristic to binarize task vectors. Specifically, we use pulse activation to remove redundant parameters, binarize the remaining non-zero parameters, and scale them back to the full-precision task vector's length. Further controlled experiments show that this binarization approximation significantly reduces the storage burden with almost no decrease in fine-tuning and merging performance. In fact, as the proportion of discarded redundant parameters increases, performance improvements are even more pronounced.

Based on these findings, we propose Task Switch (T-Switch), an efficient dynamic merging method. By utilizing pulse activation to generate a binary approximation of task vectors, we construct a "Task Switch" consisting of three components: 1) a "Activation Switch" formed by a binary mask matrix, 2) a "Polarity Switch" created from a binary sign matrix, and 3) a "Switch Knob" instantiated by a scaling factor. Building on this, we use a shared all-ones matrix to enable dynamic parameter switching across different tasks. To further automate the switch combination, we introduce Auto-Switch, which constructs a query set from the features of a small example set and performs training-free, automated switch weight allocation through retrieval on the query set. Experimental results across a range of visual and language tasks demonstrate that, compared to SOTA baselines, T-Switch and Auto-Switch achieve significant performance improvements while requiring only 1-3$\%$ of the storage space of full-precision parameters.

Our contributions can be summarized as follows:

\begin{itemize}
    \item Through controlled experiments, we get an impressive observation: task vectors exhibit a pulse-like characteristic, where parameters with smaller magnitudes are typically redundant, and discarding them simultaneously improves the performance of both fine-tuned and merged models.
    \item Based on this pulse-like characteristic, we propose T-Switch, an efficient dynamic merging approach that enables flexible reorganization of binarized task vectors. We further extend T-Switch to Auto-Switch, which allows for automatic switch combination through retrieval from a small query set.
    \item Experiments on a range of visual and language tasks demonstrate that our method achieve significant performance improvements, requiring only 1-3$\%$ of the storage space compared to full-precision task vector storage. Additionally, our approach also demonstrates excellent performance on models fine-tuned with LoRA.
\end{itemize}

\section{Related Works}
\label{sec:formatting}

\begin{figure*} [t]
\vspace{-10pt}
	\centering
\includegraphics[width=0.83\textwidth]{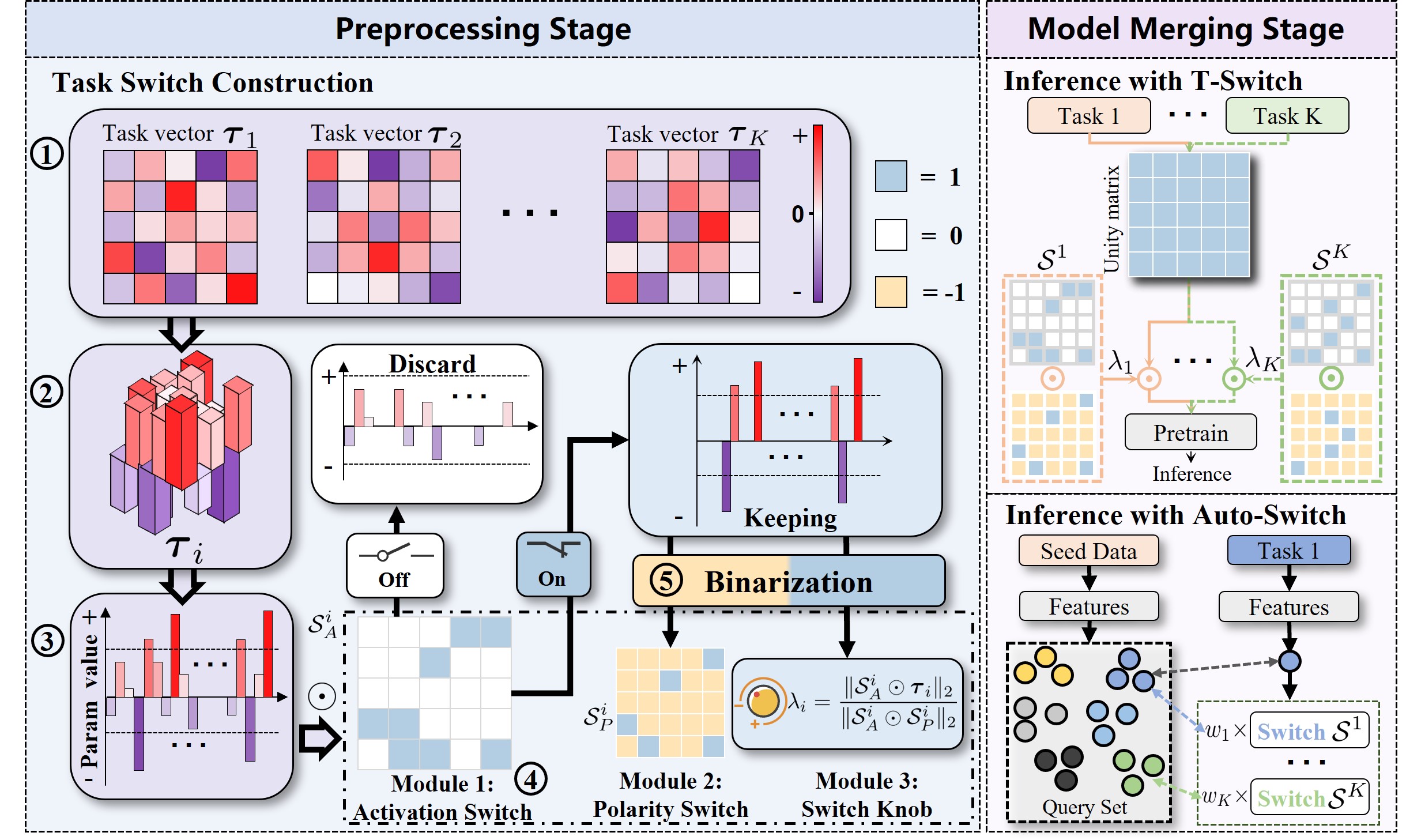}
    \vspace{-5pt}
	\caption{Overview of our method: T-Switch and Auto-Switch. The left side illustrates the construction process of the task switch, where noise parameters in the task vectors are discarded, and the remaining parameters are binarized to form the task switch. The upper right corner shows the inference process of our T-Switch using the task switch. The lower right corner demonstrates how our Auto-Switch automatically selects the task switch based on data features.}
	\label{T_Switch} 
    \vspace{-15pt}
\end{figure*}
\textbf{Model Merging.}

Model merging \cite{DBLP:conf/nips/MatenaR22,DBLP:conf/iclr/AinsworthHS23,DBLP:conf/iclr/JordanSSEN23} offers an efficient, low-cost solution for multi-task scenarios by merging multiple models fine-tuned on downstream tasks. Unlike continual learning \cite{DBLP:conf/iclr/MagistriTS0B24,DBLP:conf/cvpr/He24,DBLP:conf/iccv/ZhengMWQYY23,DBLP:journals/corr/abs-2403-19137,DBLP:conf/iclr/KimHSM23}, which required repeated fine-tuning, model merging avoids additional training and computational overhead while preserving performance on previous tasks. The simplest merging method is weight averaging \cite{DBLP:conf/icml/WortsmanIGRLMNF22}, but this often results in significant performance degradation.
Task Arithmetic \cite{DBLP:conf/iclr/IlharcoRWSHF23} introduces task vectors by calculating the difference between the weights of fine-tuned model and the pre-trained model, making merging operations more manageable. However, this method remains a linear operation on parameters, limiting its ability to preserve the multi-task capabilities of fine-tuned models. A major issue with these weight interpolation-based merging methods is parameter interference: many redundant parameter values across task-specific models conflict at certain positions, leading to adverse effects in simple interpolation.
TIES-Merging \cite{DBLP:conf/nips/YadavTCRB23} addresses this issue through three steps: resetting minimal parameter, resolving sign conflicts, and merging only aligned parameters. However, its performance relies heavily on manually set merging coefficients. Building on this, AdaMerging \cite{DBLP:conf/iclr/YangW00G0T24} introduces adaptive learning to automatically obtain merging coefficients, but this requires additional training, adding computational burden. DARE \cite{DBLP:conf/icml/Yu0Y0L24} alleviates parameter interference by randomly dropping parameters and scaling task vectors, but the performance gains from this random dropping strategy are limited.
Additionally, other methods, such as Fisher-Merging \cite{DBLP:conf/nips/MatenaR22} and RegMean \cite{DBLP:conf/iclr/Jin0P023}, leverage fisher information matrices and inner product matrices, respectively, to calculate merging coefficients for model merging. These methods often involve complex gradient calculations, which not only increase computational burden but also introduce instability issues.

\noindent\textbf{Binarization Techniques.}
Deep neural networks commonly suffer from high computational costs and memory usage. Model quantization, especially binarization, emerges as an effective way to address these issues. Binarization compresses parameters into binary bits, enabling efficient XNOR and bit-count operations that save memory, reduce energy consumption, and accelerate computation. Binarization techniques are initially focused on Convolutional Neural Networks (CNNs).
BinaryConnect \cite{DBLP:conf/nips/CourbariauxBD15} first demonstrates the feasibility of binarization on CIFAR-10 \cite{krizhevsky2009learning}. Later, XNOR-Net \cite{DBLP:conf/eccv/RastegariORF16} introduces real-valued scaling factors to improve memory and computation efficiency. To mitigate accuracy loss, Bi-Real Net \cite{DBLP:conf/eccv/LiuWLYLC18} preserves real-valued downsampling layers, addressing skip connection signal issues. 
XNOR-Net++ \cite{DBLP:conf/bmvc/BulatT19} further optimizes accuracy by combining scaling factors for activations and weights. Building on these advances, BiPer \cite{vargas2024biper} uses a binary periodic function to improve binary neural network performance, and A\&B BNN \cite{Ma_2024_CVPR} addresses hardware inefficiencies by replacing full-precision multiplications with efficient bit operations using a mask layer and a quantized RPReLU structure. Currently, binarized neural networks are mainly applied in computer vision \cite{DBLP:conf/nips/CourbariauxBD15,DBLP:journals/corr/ZhouNZWWZ16,DBLP:conf/eccv/ZhangYYH18,DBLP:conf/eccv/LiuWLYLC18,DBLP:conf/iccv/GongLJLHLYY19,DBLP:journals/caaitrit/SunYWXWG18,DBLP:conf/cvpr/LiWLQYF19}, but as their potential for resource savings becomes more apparent, the application of binarization techniques is expanding into other fields like natural language processing \cite{DBLP:conf/interspeech/XiangQ017,DBLP:journals/jzusc/QianX19,gao20211} and pattern recognition \cite{DBLP:journals/ijon/QiaoHCRNYL20}.

\section{Methodology}

\subsection{Problem Formulation}

Consider a set of tasks $\{\mathcal{T}_i\}_{i=1}^{K}$, where samples $(x_i, y_i) \in \mathcal{T}_i$ belongs to task $\mathcal{T}_i$, a pre-trained model ${f}_{\boldsymbol{\theta}}$ parameterized by pre-trained weights $\boldsymbol{\theta} \in \mathbb{R}^{n \times 1}$, and a series of fine-tuned models on the tasks, ${f}_{\boldsymbol{\theta}_1}, \dots, {f}_{\boldsymbol{\theta}_K}$, where $\boldsymbol{\theta}_i$ are the fine-tuned weights for task $\mathcal{T}_i$. The task vector $\boldsymbol{\tau}$ corresponding to task $\mathcal{T}_i$ is then computed as $\boldsymbol{\tau}_i=\boldsymbol{\theta}_i-\boldsymbol{\theta}$. The goal of model merging is to combine the task vectors $\{\boldsymbol{\tau}_{i}\}_{i=1}^{K}$ with the pre-trained model to obtain a merged model that performs well across all tasks, i.e.,
\begin{equation}
    \text{Minimize }\mathbb{E}_{(x,y)\in\cup_{i=1}^K\mathcal{T}_{i}}\ell\left(f_{\mathcal{M}\left(\boldsymbol{\theta},\{\boldsymbol{\tau}_{i}\}_{i=1}^K\right)}(x), y\right),
\end{equation}
where $\mathcal{M}$ represents the merging operation, which can be either linear \cite{DBLP:conf/iclr/IlharcoRWSHF23} or nonlinear, such as applying certain preprocessing to the task vectors \cite{DBLP:conf/icml/Yu0Y0L24}. Additionally, it can be data-dependent, for example, by training a router \cite{DBLP:journals/corr/abs-2406-15479} to automatically weight and merge task parameters based on the input data.

\subsection{
Pulse-Like Characteristics of Task Vectors}
Previous work shows that task vectors contain a large number of redundant parameters, leading to parameter conflicts between tasks during merging \cite{DBLP:conf/nips/YadavTCRB23}. Intuitively, parameters that exhibit significant changes after fine-tuning are likely to contribute more to the task, while small fluctuations may be noise caused by factors such as improper labeling or outliers, which appear as small magnitudes in the task vectors. Therefore, we aim to investigate the relationship between parameter redundancy and its magnitude to shed light on this issue. To this end, we design the following pulse activation to discard parameters in the task vector across different magnitude ranges:
\begin{equation}
    g_{m}(\boldsymbol{\tau}_{i,j})=\left\{\begin{matrix}
 1, & \text{if } \boldsymbol{\tau}_{i,j}>\gamma_{u} \text{ or } \boldsymbol{\tau}_{i,j}<\gamma_{l} \\
  0 & \text{else} 
\end{matrix}\right.,
\end{equation}
where $\gamma_{u}$ and $\gamma_{l}$ represent the upper and lower activation levels of the impulse activation function $g_\gamma$, respectively. $\boldsymbol{\tau}_{i,j}$ denotes the $j$-th element of the task vector $\boldsymbol{\tau}_{i}$. To validate the above hypothesis, we design a control experiment using CLIP-ViT-B/32 (ViT-B/32) \cite{DBLP:conf/icml/RadfordKHRGASAM21} as the backbone model, and conduct experiments on a multi-task benchmark consisting of 8 visual tasks, including datasets: SUN397 \cite{DBLP:conf/cvpr/XiaoHEOT10}, Cars \cite{DBLP:conf/iccvw/Krause0DF13}, RESISC45 \cite{DBLP:journals/pieee/ChengHL17}, EuroSAT \cite{DBLP:journals/staeors/HelberBDB19}, SVHN \cite{netzer2011reading}, GTSRB \cite{DBLP:conf/ijcnn/StallkampSSI11}, MNIST \cite{DBLP:MNIST}, and DTD \cite{DBLP:conf/cvpr/CimpoiMKMV14}. We set up the following four control conditions to provide comparative support:  
1) $\gamma_u$ and $\gamma_l$ are selected as the $\alpha$-quantiles of the positive and negative elements in $\boldsymbol{\tau}_{i}$, respectively, to simultaneously discard $\alpha$-proportion of positive and $\alpha$-proportion of negative elements with the smallest magnitude, i.e., $\boldsymbol{\tau}_{i}^\gamma = \boldsymbol{\tau}_{i} \odot  g_{m}(\boldsymbol{\tau}_{i})$; 2) Positive and negative task vector parameters with magnitudes higher than $\gamma_u$ and $\gamma_l$ are discarded, i.e., $\boldsymbol{\tau}_{i}^\gamma = \boldsymbol{\tau}_{i} \odot  (1 - g_{m})(\boldsymbol{\tau}_{i})$. This is to verify the contribution differences between high-magnitude and low-magnitude parameters; 3) Use the same random discard-and-scale strategy as DARE to randomly discard $\alpha$ proportion of task vector parameters and scale the remaining parameters by $\frac{1}{1-\alpha}$, serving as a zero control with no specific discard strategy.
\begin{figure}[t]
\vspace{-8pt}
    \centering
    \begin{minipage}{0.232\textwidth} 
        \centering
        \includegraphics[width=\textwidth]{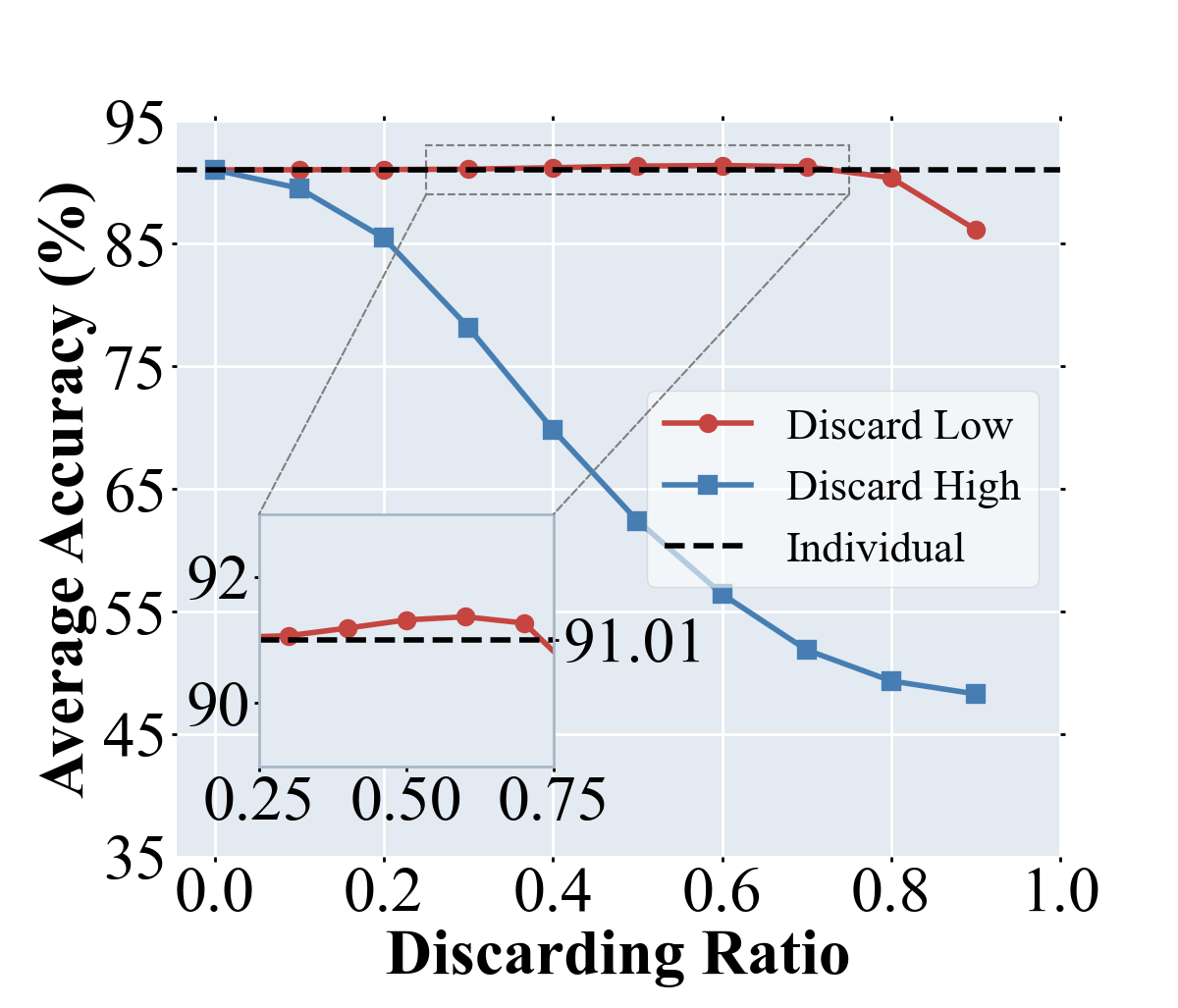}
    \end{minipage}%
    \begin{minipage}{0.232\textwidth}
        \centering
        \includegraphics[width=\textwidth]{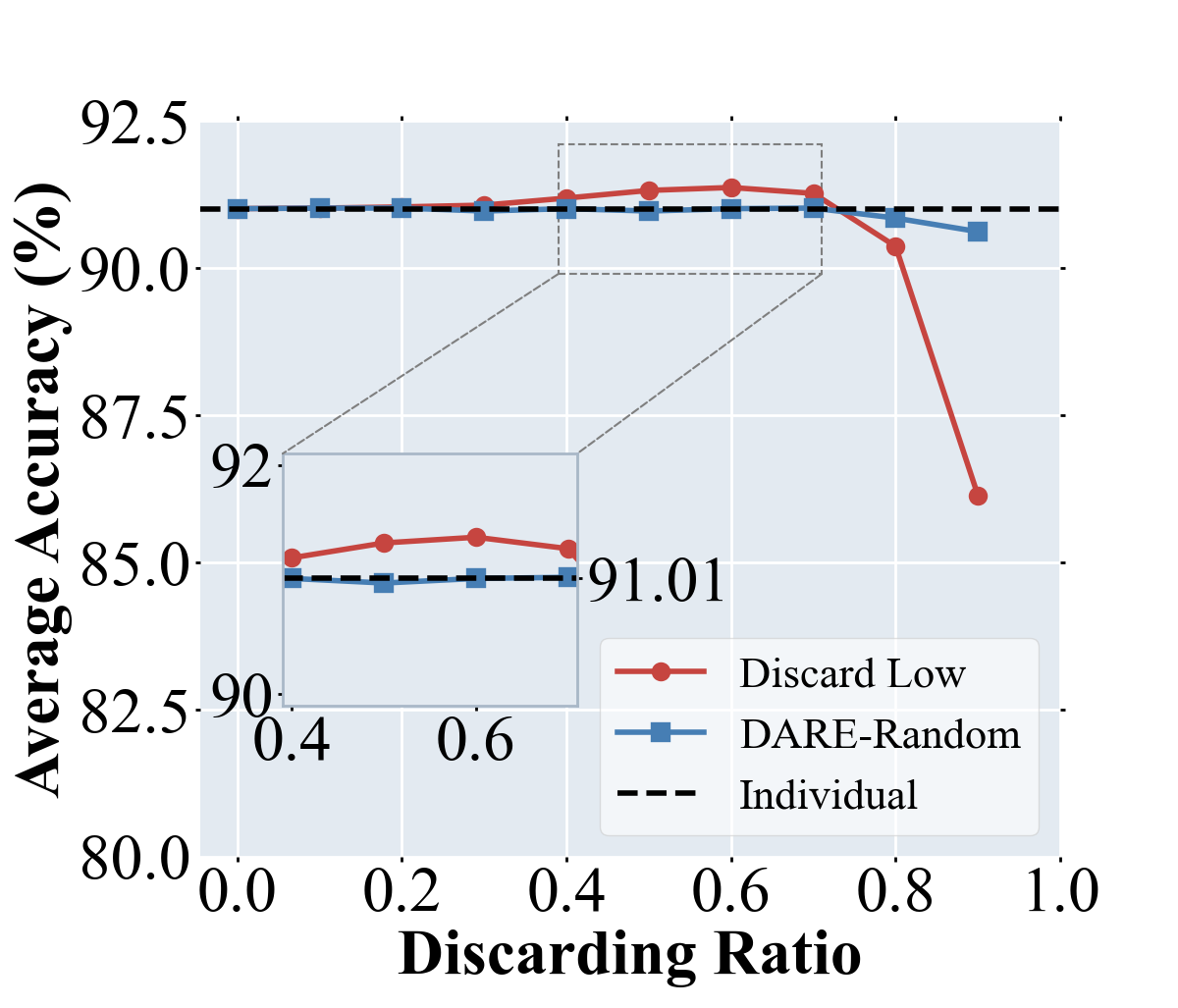}
    \end{minipage}%
    \vspace{-5pt}
    \caption{Left: comparison of the performance when discarding from smallest to largest versus from largest to smallest. Right: comparison between discarding from smallest to largest and DARE's random discarding.}
    \label{fig:Exp_3.2}
    \vspace{-15pt}
\end{figure}

The results shown in Fig.\ref{fig:Exp_3.2} indicate that as the proportion of discarded low-magnitude elements (Discard Low) increases, the average performance across all tasks does not show a significant decline. In fact, it gradually improves, and even surpasses the average performance of individual fine-tuned models (Individual) before discarding. Only after a discard rate of $\alpha = 0.7$ does the average performance begin to show a slight degradation compared to the Individual case. In contrast, discarding high-magnitude task vector parameters (Discard High) leads to a noticeable performance drop right from the start, and this decline accelerates quickly, confirming that high-magnitude task vector parameters contribute more significantly to the task. Furthermore, the strategy of DARE (DARE-Random), which uses random discard and scaling, does not exhibit performance improvement as the discard rate increases. This suggests that low-magnitude task vector parameters not only contribute little to the task, but also impose constraints on the fine-tuned model's performance. Discarding these low-magnitude parameters helps to remove redundancy and further alleviate this constraint.

These comparative results strongly confirm that task vectors exhibit pulse activation characteristics. By leveraging this property to discard low-magnitude parameters, we can not only alleviate parameter redundancy but also improve task performance. Based on this obervation, we use the following Pulse Discard (P-Discard) $g_{p}^\alpha$ to eliminate redundant parameters in task vectors:
\begin{equation}
    g_{p}^\alpha(\boldsymbol{\tau}_i) = g_{m}(\boldsymbol{\tau}_i) \odot \boldsymbol{\tau}_i,
\end{equation}

where $\alpha$ is a hyperparameter that represents the proportion of redundant parameters to be discarded. Furthermore, we aim to explore whether P-Discard can better mitigate conflicts between task vectors, thereby enhancing performance after merging. To this end, we directly merge the task vectors $\{\boldsymbol{\tau}_{i}^\alpha\}_{i=1}^K$ obtained by P-Discard, and the task vectors $\{\boldsymbol{\tau}_{i}^{\text{DARE}}\}_{i=1}^K$ obtained via random discarding with DARE at the same ratio $\alpha$, to compare the model's performance in both merging cases. Specifically, we use the following merging scheme to eliminate the effects of the discrepancy in merged vector lengths:
\begin{equation}
    \mathcal{M}\left(\boldsymbol{\theta},\{\boldsymbol{\tau}_{i}\}_{i=1}^K\right)=\boldsymbol{\theta}+\sum_{i=1}^{K}\frac{\sum_{i=1}^{K}\left\|\boldsymbol{\tau}_i\right\|_2}{\|\sum_{i=1}^{K}\boldsymbol{\tau}_i\|_2}*\boldsymbol{\tau}_i.
    \label{direct_merge}
\end{equation}

\begin{table}[t]
    \centering
    \renewcommand{\arraystretch}{1.0}
    \resizebox{0.45\textwidth}{!}{\begin{tabular}{c c c c c c c c c c}
        \toprule     
        \rowcolor{gray!30}
        Discard Ratio $\alpha$ & 0.1 & 0.2 & 0.3 & 0.4 & 0.5 & 0.6 & 0.7 & 0.8 & 0.9 \\
        \hline
        DARE-Random & 69.06 & 68.81 & 68.53 & 68.06 & 67.77 & 67.15 & 66.56 & 66.09 & 64.52 \\
        P-Discard & 69.31 & 69.45 & 69.78 & 70.41 & 71.15 & 71.95 & 72.23 & 70.99 & 66.08 \\
        \bottomrule
    \end{tabular}}
    \vspace{-5pt}
    \caption{Average merging results on eight vision datasets under different discard ratios, with the ViT-B/32 as backbone.}
    \label{toy-average-merging}
    \vspace{-10pt}
\end{table}
The results in Table \ref{toy-average-merging} indicate that compared to random discarding, P-Discard leads to a more significant improvement in merging performance. Notably, as the discarding ratio $\alpha$ increases, P-Discard continues to exhibit performance growth, whereas random discarding shows a decline in merging performance from the outset. This suggests that our P-Discard not only further improves fine-tuning performance but also alleviates conflicts between task vectors, leading to sustained improvements in merging performance.

\begin{figure}[t]
\vspace{-5pt}
    \centering
    \begin{minipage}{0.232\textwidth}
        \centering
        \includegraphics[width=\textwidth]{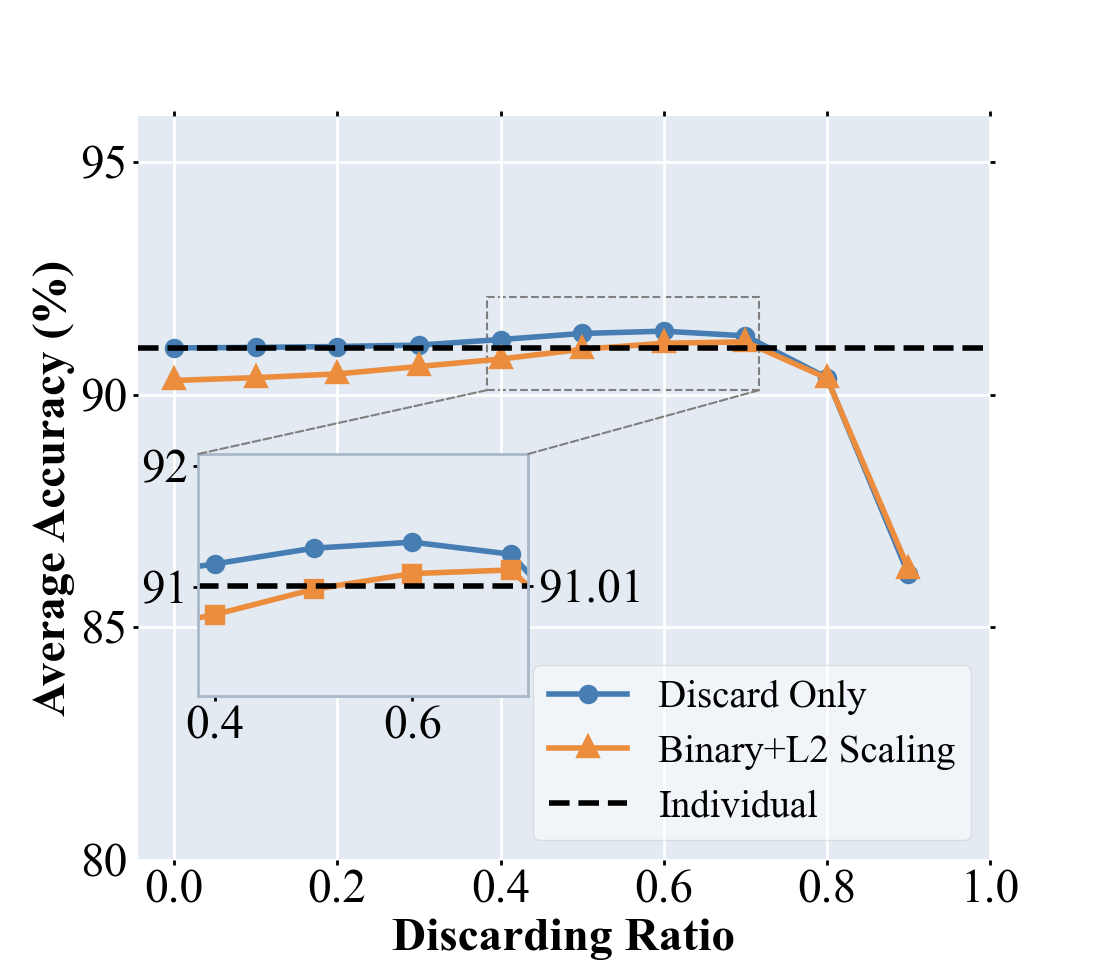}
    \end{minipage}
    \begin{minipage}{0.232\textwidth}
        \centering
        \includegraphics[width=\textwidth]{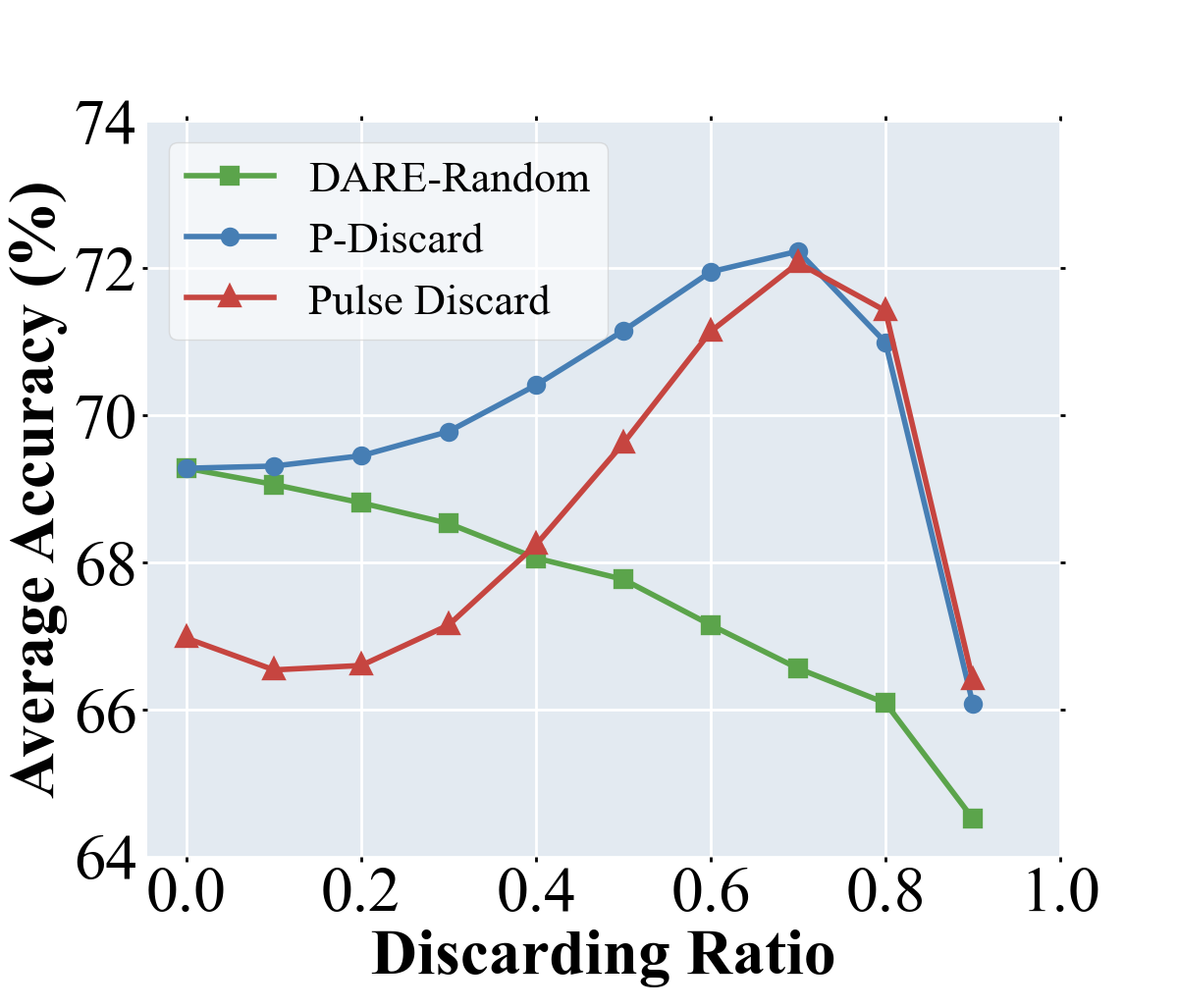}
    \end{minipage}
    \vspace{-10pt}
    \caption{Left: The average task performance of using P-Discard and Bin-Discard. Right: Comparison of the results after applying average merging and scaling on task vectors processed by DARE, P-Discard, and Bin-Discard.}
    \vspace{-3pt}
    \label{fig:Exp_3.3}
    \vspace{-15pt}
\end{figure}

\subsection{Binary Approximation of Task Vectors}

To further reduce the storage burden of parameters after P-Discard, we extend P-Discard to Binary Discard (Bin-Discard) based on the pulse activation characteristics of task vectors. Specifically, we consider the following approximation to binarize the task vector $\boldsymbol{\tau}_i$.
\begin{equation}
\hat{\boldsymbol{\tau}}_{i}=\frac{\|\boldsymbol{\tau}_{i}\odot g_{m}(\boldsymbol{\tau}_{i})\|_2}{\|g_{m}(\boldsymbol{\tau}_{i})\odot g_{b}(\boldsymbol{\tau}_{i}) \|_2}* g_{m}(\boldsymbol{\tau}_{i})\odot g_{b}(\boldsymbol{\tau}_{i}),
\label{bin:approx}
\end{equation}
where
\begin{equation}
    g_{b}(\boldsymbol{\tau}_{i,j})=\left\{\begin{matrix}
 1, & \text{if } \boldsymbol{\tau}_{i,j}>0 \\
 -1, & \text{if }\boldsymbol{\tau}_{i,j}\le 0\\
\end{matrix}\right..
\end{equation}

This approximation suggests that we can approximate the original task vector $\boldsymbol{\tau}_{i}$ using only a binary mask matrix $g_{m}(\boldsymbol{\tau}_{i})$ for removing redundancies, a binary sign matrix $g_{b}(\boldsymbol{\tau}_{i})$ that captures the direction, and a scaling factor $\frac{\|\boldsymbol{\tau}_{i}\odot g_{m}(\boldsymbol{\tau}_{i})\|_2}{\|g_{m}(\boldsymbol{\tau}_{i})\odot g_{b}(\boldsymbol{\tau}_{i}) \|_2}$. If this approximation can retain most of the performance of the original task vector, it could greatly improve the storage efficiency of task vectors while eliminating parameter redundancy. 
\begin{figure}[t]
\vspace{-5pt}
    \centering
    \includegraphics[width=0.48\textwidth]{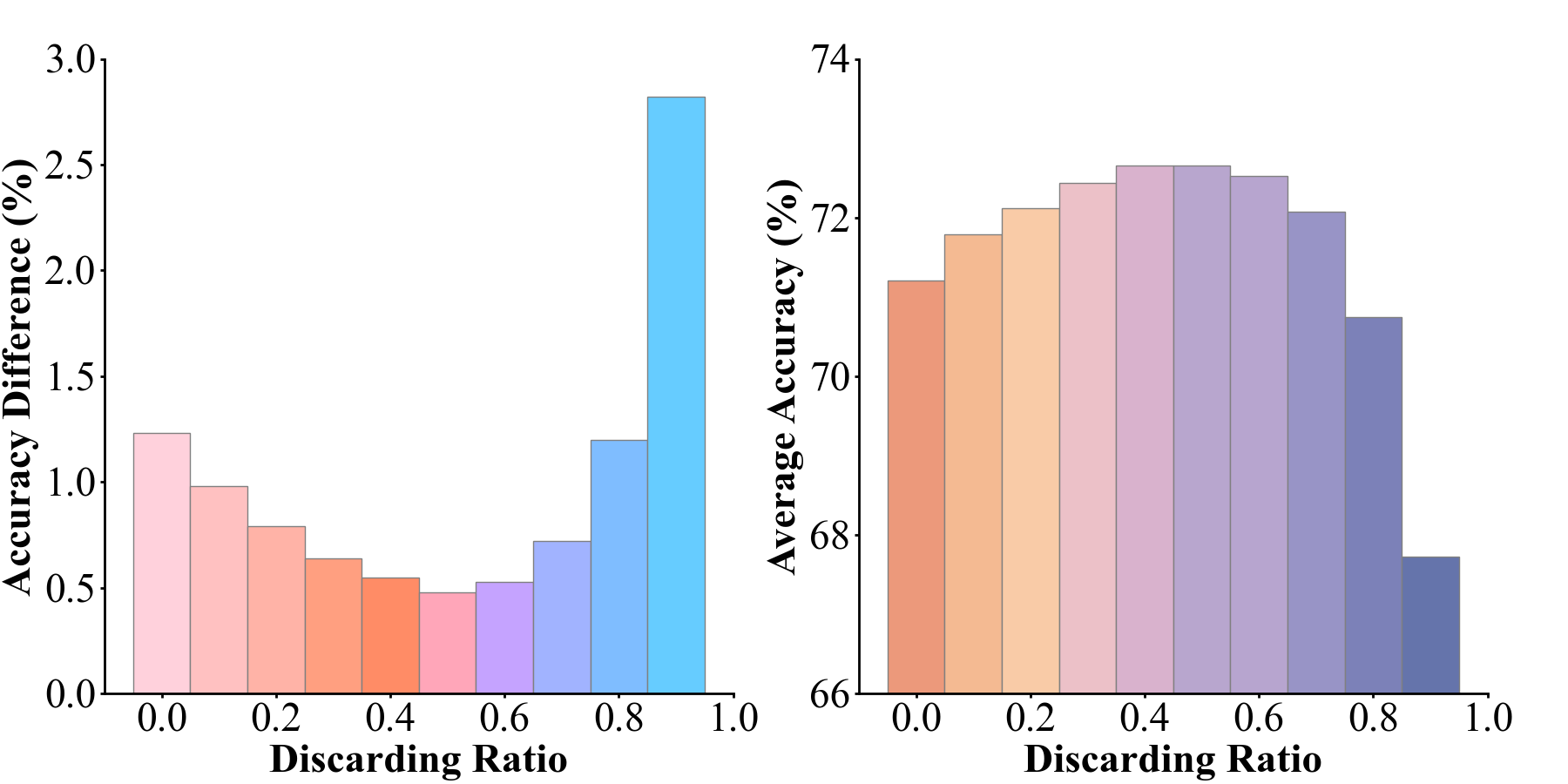}  
    \caption{Left: The accuracy difference
between Bin-Discard and individual fine-tuning performance as the discard rate varies. Right: The results of merging after applying the Bin-Discard as the discard rate varies.}  
    \label{fig:lora_merge}  
    \vspace{-15pt}
\end{figure}

To verify the feasibility of this approximation, we apply Bin-Discard to binarize task vector parameters at different discard ratios, and test the performance of the binarized task vectors on their corresponding tasks under the experimental setup of the previous section. The results, shown in the left panel of Fig. \ref{fig:Exp_3.3}, indicate that even in this binarized approximation, the performance of the task vectors hardly suffers any degradation. Moreover, as the discard rate increases, the performance of the binarized approximated task vectors gradually approaches that of the full-precision task vectors at the corresponding discard rate, both demonstrating a performance enhancement. Surprisingly, when the discard rate is between $\alpha=0.6$ and $\alpha=0.7$, the performance of the binarized approximated task vectors even surpasses that of the original full-precision fine-tuned models. This suggests that task vector binarization is not only feasible, but also benefits from the removal of redundant parameters, leading to a stronger performance gain. 

To further confirm whether Bin-Discard can maintain the performance of full-precision task vectors in model merging, we conducted experiments using the same experimental setup as in the previous section, but with task vectors binarized by Bin-Discard. The results shown in the right panel of Fig.\ref{fig:Exp_3.3} demonstrate that even under binarized approximation, an excellent merging performance is achieved, with a performance improvement as the discard rate increases. Additionally, considering that parameter-efficient fine-tuning methods, such as LoRA, have become a mainstream paradigm in fine-tuning practices, we also apply this binarized approximation to a group of fine-tuned LoRA vectors, and conducted the same merging experiments. Results shown in Fig.\ref{fig:lora_merge} indicate that binarized task vector approximation also helps in removing redundancies in low-rank task vectors and improving model merging. These findings strongly support the conclusion that task vector binarization has the potential to alleviate task vector conflicts while significantly reducing storage requirements.

\subsection{Dynamic Merging with Binary Task Vectors}
\noindent\textbf{T-Switch for Dynamic Merging.} Based on the findings and insights from the previous sections, we propose the binary approximation-based model merging method, T-Switch, as illustrated in Fig.\ref{T_Switch}. Specifically, for each task vector $\boldsymbol{\tau}i$, we first apply Bin-Discard to binarize it, efficiently decomposing it into the following three components: 1) Activation Switch $\mathcal{S}_{A}^{i} = g_{m}(\boldsymbol{\tau}_{i})$, which activates the task vector parameters that contribute to task $\mathcal{T}_i$; 2) Polarity Switch $\mathcal{S}_{P}^{i} = g_{b}(\boldsymbol{\tau}_{i})$, which represents the direction of the task vector corresponding to task $\mathcal{T}_i$; and 3) Switch Knob $\lambda_i = \frac{\|\mathcal{S}_{A}^i \odot \boldsymbol{\tau}_i\|_2}{\|\mathcal{S}_{A}^i \odot \mathcal{S}_{P}^i\|_2}$, which provides an approximate scaling of the binarized task vector relative to the full-precision task vector. Using the task switch group $\mathcal{S}^{i}=\{\mathcal{S}_{A}^i,\mathcal{S}_{P}^i,\lambda_i\}$ formed by these three components, we dynamically apply the following merging scheme during the inference phase for each task $\mathcal{T}_i$:
\begin{equation}
    \hat{\boldsymbol{\theta}_i}=\boldsymbol{\theta}+\lambda_i*\mathcal{S}_{A}^i\odot\mathcal{S}_{P}^i\odot\mathbf{U},
\end{equation}
where $\mathbf{U} \in \mathbb{R}^{n \times 1}$ is a shared vector where all elements are 1. This indicates that for each task $\mathcal{T}_i$, T-Switch can flexibly "activate" the parameters associated with that task from the shared all-ones vector $\mathbf{U} \in \mathbb{R}^{n \times 1}$, and obtain a good approximation of the original full-precision task vector $\hat{\boldsymbol{\theta}_i}$, while ensuring highly efficient storage of the task vectors.

\noindent{\textbf{Auto-Switch for Automatic Merging}.}
Due to the fact that tasks in real-world applications may change based on the user's specific problems, we aim to further enable T-Switch to automatically switch and reassemble task vectors for different tasks. To this end, we propose an automated version of T-Switch, Auto-Switch. Considering that training a learnable router based on example data introduces additional training costs, and that the router would need to be retrained whenever a new model merging requirement arises, we avoid this inconvenience by not using a learnable parametric router. Instead, we turn to an automated switch-based combination mechanism that queries during inference. 

Specifically, we first construct a query set $\mathcal{Q}_i = \{f_{\bar{\boldsymbol{\theta}}}(x_i) \mid (x_i, y_i) \in \mathcal{E}_i\}$ for each task $\mathcal{T}_i$ using a small subset of example data $\mathcal{E}_i = \{(x_i, y_i)\}_{i=1}^N \subset \mathcal{T}_i$, where $\bar{\boldsymbol{\theta}}$ represents the model weights obtained after merging the directly averaged task vectors according to the scheme in equation \ref{direct_merge}, which helps improve the task distinguishability of the backbone model. The operation $f^{\text{ex}}$ captures the feature outputs of the model prior to the linear classifier. Note that since the query set only requires input examples, no label information is needed. Based on the constructed query sets, for each input $x$, we perform a nearest-neighbor search within the overall query set $\mathcal{Q} = \cup_{i=1}^K \mathcal{Q}_i$ to find the $C$ nearest neighbors to $f_{\bar{\boldsymbol{\theta}}}(x)$. This set of neighbors, denoted as $\mathcal{N}_x$, is then used to automatically assign weights to the task switches and execute the model merging:
\begin{equation}
    \hat{\boldsymbol{\theta}}(x)=\boldsymbol{\theta}+\sum_{i=1}^K\lambda_i w_{i}(x)*\mathcal{S}_{A}^i\odot\mathcal{S}_{P}^i\odot\mathbf{U},
\end{equation}
where $ w_i(x) = \frac{|\mathcal{Q}_i \cap \mathcal{N}_x|}{|\mathcal{N}_x|} $ represents the switch weight assigned to task $ \mathcal{T}_i $, and `$ |\cdot| $' denotes the number of elements in a set. Since Auto-Switch does not require an explicitly parameterized router, it provides greater flexibility and eliminates the need for additional training. Moreover, benefiting from the binarized task vector, it significantly outperforms conventional router-based automatic merging methods in terms of storage efficiency.
\definecolor{color1}{HTML}{F7E6E1} 
\definecolor{color2}{HTML}{E0F1F7}
\section{Experiments}
In this section, we conduct a comprehensive comparison of our T-Switch and Auto-Switch with multiple baseline methods, including model merging experiments on both vision and language models. Additionally, we conduct merging performance comparisons on task vectors fine-tuned with LoRA to thoroughly validate the advantages of our method. More experimental results and detailed experimental setups can be found in the Appendix.
\begin{table*}[h!]
\renewcommand{\arraystretch}{0.8}
    \centering
    \renewcommand{\arraystretch}{1.0}
    \resizebox{0.9\textwidth}{!}{\begin{tabular}{c l c c c c c c c c c c c c}
        \toprule
        \rowcolor{blue!15}
        Type&Methods& Automatic & Example & Storage(MB) &SUN397&Cars&RESISC45&EuroSAT&SVHN&GTSRB&MNIST&DTD&AVG \\
        \hline
        &\cellcolor{color1}Pre-trained&\cellcolor{color1}--&\cellcolor{color1} --&\cellcolor{color1} --&\cellcolor{color1}62.32&\cellcolor{color1}59.63&\cellcolor{color1}60.27&\cellcolor{color1}45.74&\cellcolor{color1}31.63&\cellcolor{color1}32.60&\cellcolor{color1}48.26&\cellcolor{color1}44.41&\cellcolor{color1}48.11 \\
        &\cellcolor{color2}Individual&\cellcolor{color2}--& \cellcolor{color2}--& \cellcolor{color2}--&\cellcolor{color2}79.23&\cellcolor{color2}77.68&\cellcolor{color2}96.11&\cellcolor{color2}99.78&\cellcolor{color2}97.46&\cellcolor{color2}98.73&\cellcolor{color2}99.69&\cellcolor{color2}79.41&\cellcolor{color2}91.01 \\
        \multirow{-3}{*}{--}&\cellcolor{color1}Traditional MTL&\cellcolor{color1}--& \cellcolor{color1}--&\cellcolor{color1} --&\cellcolor{color1}73.90&\cellcolor{color1}74.40&\cellcolor{color1}93.90&\cellcolor{color1}98.20&\cellcolor{color1}95.80&\cellcolor{color1}98.90&\cellcolor{color1}99.50&\cellcolor{color1}77.90&\cellcolor{color1}89.06 \\
        \hline
        &\cellcolor{color2}Weight-Averaging&\cellcolor{color2}--&\cellcolor{color2}\textcolor{blue!70}{\faTimes}&\cellcolor{color2}-- &\cellcolor{color2}64.72&\cellcolor{color2}63.34&\cellcolor{color2}71.46&\cellcolor{color2}72.74&\cellcolor{color2}64.16&\cellcolor{color2}52.79&\cellcolor{color2}87.46&\cellcolor{color2}50.11&\cellcolor{color2}65.85 \\
        &\cellcolor{color1}Task-Arithmetic&\cellcolor{color1}--&\cellcolor{color1}\textcolor{blue!70}{\faTimes}&\cellcolor{color1}--&\cellcolor{color1}63.50&\cellcolor{color1}62.04&\cellcolor{color1}72.00&\cellcolor{color1}78.59&\cellcolor{color1}74.43&\cellcolor{color1}65.09&\cellcolor{color1}94.00&\cellcolor{color1}52.18&\cellcolor{color1}70.23 \\
        &\cellcolor{color2}Ties-Merging&\cellcolor{color2}--&\cellcolor{color2}\textcolor{blue!70}{\faTimes}&\cellcolor{color2}--&\cellcolor{color2}64.99&\cellcolor{color2}64.30&\cellcolor{color2}74.65&\cellcolor{color2}76.48&\cellcolor{color2}81.28&\cellcolor{color2}69.38&\cellcolor{color2}96.53&\cellcolor{color2}54.26&\cellcolor{color2}72.73 \\
        &\cellcolor{color1}DARE&\cellcolor{color1}--&\cellcolor{color1}\textcolor{blue!70}{\faTimes}&\cellcolor{color1}--&\cellcolor{color1}64.76&\cellcolor{color1}63.08&\cellcolor{color1}71.02&\cellcolor{color1}70.70&\cellcolor{color1}62.04&\cellcolor{color1}50.68&\cellcolor{color1}86.17&\cellcolor{color1}50.64&\cellcolor{color1}64.89 \\
        &\cellcolor{color2}RegMean&\cellcolor{color2}--& \cellcolor{color2}\textcolor{red!70}{\faCheck}&\cellcolor{color2}--&\cellcolor{color2}66.96&\cellcolor{color2}65.84&\cellcolor{color2}80.92&\cellcolor{color2}91.74&\cellcolor{color2}84.90&\cellcolor{color2}78.43&\cellcolor{color2}96.49&\cellcolor{color2}60.53&\cellcolor{color2}78.23 \\
        &\cellcolor{color2}Fisher Merging&\cellcolor{color2}--& \cellcolor{color2}\textcolor{red!70}{\faCheck}&\cellcolor{color2}--&\cellcolor{color2}67.13&\cellcolor{color2}66.72&\cellcolor{color2}71.68&\cellcolor{color2}64.07&\cellcolor{color2}85.04&\cellcolor{color2}72.47&\cellcolor{color2}85.59&\cellcolor{color2}51.01&\cellcolor{color2}70.46 \\
        &\cellcolor{color1}AdaMerging&\cellcolor{color1}--&\cellcolor{color1}\textcolor{red!70}{\faCheck}&\cellcolor{color1}--&\cellcolor{color1}64.51&\cellcolor{color1}67.90&\cellcolor{color1}79.73&\cellcolor{color1}93.19&\cellcolor{color1}86.31&\cellcolor{color1}92.36&\cellcolor{color1}97.53&\cellcolor{color1}58.62&\cellcolor{color1}80.02 \\
        \multirow{-8}{*}{\rotatebox{90}{\textbf{Fixed}}}&\cellcolor{color1}AdaMerging++&\cellcolor{color1}--&\cellcolor{color1}\textcolor{red!70}{\faCheck}&\cellcolor{color1}--&\cellcolor{color1}66.73&\cellcolor{color1}68.42&\cellcolor{color1}81.95&\cellcolor{color1}93.52&\cellcolor{color1}89.53&\cellcolor{color1}89.44&\cellcolor{color1}98.30&\cellcolor{color1}60.27&\cellcolor{color1}81.02 \\
        
        \midrule
        &\cellcolor{color2}Twin-merging&\cellcolor{color2}\textcolor{blue!70}{\faCheck}&\cellcolor{color2}\textcolor{red!70}{\faCheck}&\cellcolor{color2}3474.2&\cellcolor{color2}71.56&\cellcolor{color2}68.78&\cellcolor{color2}89.97&\cellcolor{color2}72.11&\cellcolor{color2}96.65&\cellcolor{color2}93.35&\cellcolor{color2}99.66&\cellcolor{color2}72.50&\cellcolor{color2}83.07 \\
        &\cellcolor{color1}EMR-merging&\cellcolor{color1}\textcolor{red!70}{\faTimes}&\cellcolor{color1}\textcolor{blue!70}{\faTimes}&\cellcolor{color1}461.0&\cellcolor{color1}75.19& \cellcolor{color1}72.76&\cellcolor{color1}93.49& \cellcolor{color1}99.52& \cellcolor{color1}96.86& \cellcolor{color1}98.13& \cellcolor{color1}99.58& \cellcolor{color1}74.36& \cellcolor{color1}88.74 \\
        &\cellcolor{color2} T-Switch (Ours) & \cellcolor{color2}\textcolor{red!70}{\faTimes} &\cellcolor{color2}\textcolor{blue!70}{\faTimes}&\cellcolor{color2}57.0&\cellcolor{color2}\textbf{79.36}& \cellcolor{color2}\underline{77.60}&\cellcolor{color2}\textbf{95.98}&\cellcolor{color2}\textbf{99.74}&\cellcolor{color2}\textbf{97.33}&\cellcolor{color2}\textbf{98.61}&\cellcolor{color2}\textbf{99.68}&\cellcolor{color2}\textbf{79.52}&\cellcolor{color2}\textbf{90.98} \\
        \multirow{-4}{*}{\rotatebox{90}{\textbf{Dynamic}}}& \cellcolor{color1}Auto-Switch (Ours) & \cellcolor{color1}\textcolor{blue!70}{\faCheck}& \cellcolor{color1}\textcolor{red!70}{\faCheck}&\cellcolor{color1}58.6&\cellcolor{color1}\underline{76.08}&\cellcolor{color1}\textbf{77.64}&\cellcolor{color1}\underline{93.60}&\cellcolor{color1}\textbf{99.74}&\cellcolor{color1}\textbf{97.33}&\cellcolor{color1}\underline{98.59}&\cellcolor{color1}\textbf{99.68}&\cellcolor{color1}\underline{79.31}&\cellcolor{color1}\underline{90.25}\\
        \bottomrule
    \end{tabular}}
    \vspace{-3pt}
    \caption{Main results of merging full-rank task vectors of the ViT-B/32 model on eight vision datasets. The best method is highlighted in bold, and the second-best method is underlined.}
    \label{ViT-B-32}
    \vspace{-2pt}
\end{table*}

\begin{table*}[h!]
\renewcommand{\arraystretch}{0.8}
    \centering
    \renewcommand{\arraystretch}{1.0}
    \resizebox{0.9\textwidth}{!}{\begin{tabular}{c l c c c c c c c c c c c c}
        \toprule
        \rowcolor{blue!15}
        Type&Methods& Automatic & Example & Storage(MB)&SUN397&Cars&RESISC45&EuroSAT&SVHN&GTSRB&MNIST&DTD&AVG \\
        \hline
        \multirow{3}{*}{--}&\cellcolor{color1}Pre-trained&\cellcolor{color1}--& \cellcolor{color1}-- & \cellcolor{color1}--&\cellcolor{color1}62.32&\cellcolor{color1}59.63&\cellcolor{color1}60.27&\cellcolor{color1}45.74&\cellcolor{color1}31.63&\cellcolor{color1}32.60&\cellcolor{color1}48.26&\cellcolor{color1}44.41&\cellcolor{color1}48.11 \\
        &\cellcolor{color2}LoRA Finetuned&\cellcolor{color2}--& \cellcolor{color2}--&\cellcolor{color2} --&\cellcolor{color2}73.10&\cellcolor{color2}66.09&\cellcolor{color2}93.71&\cellcolor{color2}98.56&\cellcolor{color2}96.93&\cellcolor{color2}98.69&\cellcolor{color2}99.62&\cellcolor{color2}65.43&\cellcolor{color2}86.52 \\
        \hline
        \multirow{8}{*}{\rotatebox{90}{\textbf{Static}}}&\cellcolor{color1}Weight-Averaging&\cellcolor{color1}--&\cellcolor{color1}\textcolor{blue!70}{\faTimes}& \cellcolor{color1}--&\cellcolor{color1}64.36&\cellcolor{color1}61.81&\cellcolor{color1}70.83&\cellcolor{color1}71.26&\cellcolor{color1}63.63&\cellcolor{color1}52.19&\cellcolor{color1}81.72&\cellcolor{color1}47.39&\cellcolor{color1}64.15 \\
        &\cellcolor{color2}Task-Arithmetic&\cellcolor{color2}--&\cellcolor{color2}\textcolor{blue!70}{\faTimes}&\cellcolor{color2} --&\cellcolor{color2}63.39&\cellcolor{color2}58.15&\cellcolor{color2}74.60&\cellcolor{color2}81.89&\cellcolor{color2}85.00&\cellcolor{color2}75.03&\cellcolor{color2}94.61&\cellcolor{color2}49.63&\cellcolor{color2}72.79 \\
        &\cellcolor{color1}Ties-Merging&\cellcolor{color1}--&\cellcolor{color1}\textcolor{blue!70}{\faTimes}&\cellcolor{color1} --&\cellcolor{color1}59.20&\cellcolor{color1}50.81&\cellcolor{color1}71.57&\cellcolor{color1}77.41&\cellcolor{color1}89.38&\cellcolor{color1}78.23&\cellcolor{color1}96.95&\cellcolor{color1}46.76&\cellcolor{color1}71.29 \\
        &\cellcolor{color2}DARE&\cellcolor{color2}--&\cellcolor{color2}\textcolor{blue!70}{\faTimes}& \cellcolor{color2}--&\cellcolor{color2}64.48&\cellcolor{color2}60.30&\cellcolor{color2}74.44&\cellcolor{color2}79.59&\cellcolor{color2}79.67&\cellcolor{color2}69.80&\cellcolor{color2}92.24&\cellcolor{color2}49.89&\cellcolor{color2}71.30 \\
        &\cellcolor{color1}RegMean&\cellcolor{color1}--& \cellcolor{color1}\textcolor{red!70}{\faCheck}& \cellcolor{color1}--&\cellcolor{color1}66.89&\cellcolor{color1}64.84&\cellcolor{color1}82.68&\cellcolor{color1}93.07&\cellcolor{color1}89.97&\cellcolor{color1}85.56&\cellcolor{color1}97.28&\cellcolor{color1}55.48&\cellcolor{color1}79.47 \\
        &\cellcolor{color2}Fisher Merging&\cellcolor{color2}--& \cellcolor{color2}\textcolor{red!70}{\faCheck}&\cellcolor{color2} --&\cellcolor{color2}66.30&\cellcolor{color2}63.95&\cellcolor{color2}77.19&\cellcolor{color2}75.85&\cellcolor{color2}81.95&\cellcolor{color2}78.25&\cellcolor{color2}81.77&\cellcolor{color2}50.80&\cellcolor{color2}72.01 \\
        &\cellcolor{color1}AdaMerging&\cellcolor{color1}--&\cellcolor{color1}\textcolor{red!70}{\faCheck}& \cellcolor{color1}--&\cellcolor{color1}63.80&\cellcolor{color1}57.42&\cellcolor{color1}78.75&\cellcolor{color1}93.04&\cellcolor{color1}79.56&\cellcolor{color1}87.70&\cellcolor{color1}96.03&\cellcolor{color1}50.11&\cellcolor{color1}75.80 \\
        &\cellcolor{color2}AdaMerging++&\cellcolor{color2}--&\cellcolor{color2}\textcolor{red!70}{\faCheck}& \cellcolor{color2}--&\cellcolor{color2}64.11&\cellcolor{color2}56.82&\cellcolor{color2}79.13&\cellcolor{color2}91.30&\cellcolor{color2}81.35&\cellcolor{color2}86.15&\cellcolor{color2}96.46&\cellcolor{color2}51.01&\cellcolor{color2}75.79 \\
        \midrule
        \multirow{4}{*}{\rotatebox{90}{\textbf{Dynamic}}}&\cellcolor{color1}Twin-merging&\cellcolor{color1}\textcolor{blue!70}{\faCheck}&\cellcolor{color1}\textcolor{red!70}{\faCheck}&\cellcolor{color1}1810.9&\cellcolor{color1}\underline{72.44}&\cellcolor{color1}64.71&\cellcolor{color1}\underline{92.32}&\cellcolor{color1}\textbf{98.52}&\cellcolor{color1}\textbf{96.84}&\cellcolor{color1}95.84&\cellcolor{color1}\textbf{99.48}&\cellcolor{color1}\textbf{66.70}&\cellcolor{color1}\underline{85.86} \\
        &\cellcolor{color2}EMR-merging&\cellcolor{color2}\textcolor{red!70}{\faTimes}&\cellcolor{color2}\textcolor{blue!70}{\faTimes}&\cellcolor{color2}239.7&\cellcolor{color2}70.46&\cellcolor{color2}64.86&\cellcolor{color2}91.11& \cellcolor{color2}97.63&\cellcolor{color2}96.11&\cellcolor{color2}97.13& \cellcolor{color2}99.30&\cellcolor{color2}59.10&\cellcolor{color2}84.46 \\
        &\cellcolor{color1} T-Switch (Ours) & \cellcolor{color1} \textcolor{red!70}{\faTimes} &\cellcolor{color1} \textcolor{blue!70}{\faTimes} & \cellcolor{color1}32.0&\cellcolor{color1} \textbf{72.66}& \cellcolor{color1} \underline{67.57}&\cellcolor{color1} \textbf{92.97}&\cellcolor{color1} \underline{98.33}&\cellcolor{color1} \underline{96.43}&\cellcolor{color1} \textbf{97.98}&\cellcolor{color1} \underline{99.47}&\cellcolor{color1} \underline{62.87}&\cellcolor{color1} \textbf{86.04} \\
        & \cellcolor{color2} Auto-Switch (Ours) & \cellcolor{color2} \textcolor{blue!70}{\faCheck}& \cellcolor{color2} \textcolor{red!70}{\faCheck}&\cellcolor{color2} 33.6&\cellcolor{color2} 70.05&\cellcolor{color2} \textbf{67.58}&\cellcolor{color2} 89.56&\cellcolor{color2} 98.26&\cellcolor{color2} 96.39&\cellcolor{color2} \underline{97.38}&\cellcolor{color2} \underline{99.47}&\cellcolor{color2} 59.95&\cellcolor{color2} 84.83\\
        \bottomrule
    \end{tabular}}
    \vspace{-3pt}
    \caption{Main results of merging low-rank task vectors of the ViT-B/32 model on eight vision datasets. The best method is highlighted in bold, and the second-best method is underlined.}
    \label{LoRA_ViT-B-32}
    \vspace{-13pt}
\end{table*}

\begin{table*}
\renewcommand{\arraystretch}{0.8}
    \centering
    \renewcommand{\arraystretch}{1.0}
    \resizebox{0.9\textwidth}{!}{\begin{tabular}{c l c c c c c c c c c c c c}
        \toprule[1.5pt]
        \rowcolor{blue!15}
        Type&Methods&Automatic & Example & Storage(MB)&CoLA &SST2&MRPC &STSB &QQP &MNLI &QNLI &RTE&AVG \\
        \hline
        \multirow{2}{*}{--}&\cellcolor{color1} Pre-trained&\cellcolor{color1} --&\cellcolor{color1}  --&\cellcolor{color1} -- &\cellcolor{color1} 0.0000&\cellcolor{color1} 0.4908&\cellcolor{color1} 0.3162&\cellcolor{color1} 0.0440&\cellcolor{color1} 0.3682&\cellcolor{color1} 0.3182&\cellcolor{color1} 0.5089&\cellcolor{color1} 0.4729&\cellcolor{color1} 0.3149 \\
        &\cellcolor{color2} Individual&\cellcolor{color2} --&\cellcolor{color2}  --&\cellcolor{color2} --&\cellcolor{color2} 0.6018 &\cellcolor{color2} 0.9404 &\cellcolor{color2} 0.8922 &\cellcolor{color2} 0.9063 &\cellcolor{color2} 0.9141 &\cellcolor{color2} 0.8720 &\cellcolor{color2} 0.9271 &\cellcolor{color2} 0.7906&\cellcolor{color2} 0.8556 \\
        \hline
        \multirow{6}{*}{\rotatebox{90}{\textbf{Static}}}&\cellcolor{color1} Weight-Averaging&\cellcolor{color1}--&\cellcolor{color1}\textcolor{blue!70}{\faTimes}&\cellcolor{color1}--&\cellcolor{color1}0.1396 &\cellcolor{color1}0.6411 &\cellcolor{color1}0.6936 &\cellcolor{color1}0.3184 &\cellcolor{color1}0.7536 &\cellcolor{color1}0.4219 &\cellcolor{color1}0.587 &\cellcolor{color1}0.5523&\cellcolor{color1}0.5134 \\
        &\cellcolor{color2}Task-Arithmetic&\cellcolor{color2}--&\cellcolor{color2}\textcolor{blue!70}{\faTimes}&\cellcolor{color2}--&\cellcolor{color2}0.1878 &\cellcolor{color2}0.8589 &\cellcolor{color2}0.7990 &\cellcolor{color2}0.7403 &\cellcolor{color2}0.8378 &\cellcolor{color2}0.5908 &\cellcolor{color2}0.6967 &\cellcolor{color2}0.6209&\cellcolor{color2}0.6665 \\
        &\cellcolor{color1}Ties-merging&\cellcolor{color1}--&\cellcolor{color1}\textcolor{blue!70}{\faTimes}&\cellcolor{color1}--&\cellcolor{color1}0.2048 &\cellcolor{color1}0.8440 &\cellcolor{color1}0.8113 &\cellcolor{color1}0.5819 &\cellcolor{color1}0.8570 &\cellcolor{color1}0.6465 &\cellcolor{color1}0.7481 &\cellcolor{color1}0.4296&\cellcolor{color1}0.6404 \\
        &\cellcolor{color2}DARE&\cellcolor{color2}--&\cellcolor{color2}\textcolor{blue!70}{\faTimes}&\cellcolor{color2}--&\cellcolor{color2}0.0804&\cellcolor{color2}0.7924&\cellcolor{color2}0.7794&\cellcolor{color2}0.3054&\cellcolor{color2}0.7935&\cellcolor{color2}0.4000&\cellcolor{color2}0.7227&\cellcolor{color2}0.6029&\cellcolor{color2}0.5596 \\
        &\cellcolor{color1}RegMean&\cellcolor{color1}--&\cellcolor{color1}\textcolor{red!70}{\faCheck}&\cellcolor{color1}--&\cellcolor{color1}0.3667 &\cellcolor{color1}0.9060 &\cellcolor{color1}0.7574 &\cellcolor{color1}0.6268 &\cellcolor{color1}0.8355 &\cellcolor{color1}0.7002 &\cellcolor{color1}0.8235 &\cellcolor{color1}0.5848&\cellcolor{color1}0.7001 \\
        &\cellcolor{color2}Fisher Merging&\cellcolor{color2}--&\cellcolor{color2}\textcolor{red!70}{\faCheck}&\cellcolor{color2}--&\cellcolor{color2}0.1875&\cellcolor{color2}0.5482&\cellcolor{color2}0.8137&\cellcolor{color2}0.7743&\cellcolor{color2}0.8313&\cellcolor{color2}0.3496&\cellcolor{color2}0.6745&\cellcolor{color2}0.5162&\cellcolor{color2}0.5869 \\
        \midrule
        \multirow{4}{*}{\rotatebox{90}{\textbf{Dynamic}}}&\cellcolor{color1}Twin-merging&\cellcolor{color1}\textcolor{blue!70}{\faCheck}& \cellcolor{color1}\textcolor{red!70}{\faCheck}&\cellcolor{color1}3819.9&\cellcolor{color1}\textbf{0.5931}&\cellcolor{color1}0.9381&\cellcolor{color1}\textbf{0.8924}&\cellcolor{color1}0.6833&\cellcolor{color1}0.8890&\cellcolor{color1}0.8210&\cellcolor{color1}0.9030&\cellcolor{color1}0.7617&\cellcolor{color1}0.8102 \\
        &\cellcolor{color2}EMR-merging&\cellcolor{color2}\textcolor{red!70}{\faTimes}&\cellcolor{color2}\textcolor{blue!70}{\faTimes}&\cellcolor{color2}506.8&\cellcolor{color2}0.3996&\cellcolor{color2}0.9335&\cellcolor{color2}0.8627&\cellcolor{color2}0.8277&\cellcolor{color2}0.8972&\cellcolor{color2}0.8545&\cellcolor{color2}0.8957&\cellcolor{color2}0.7437&\cellcolor{color2}0.8018 \\
        &\cellcolor{color1}T-Switch (Ours) & \cellcolor{color1}\textcolor{red!70}{\faTimes}&\cellcolor{color1}\textcolor{blue!70}{\faTimes}&\cellcolor{color1}63.2&\cellcolor{color1}\underline{0.5339}&\cellcolor{color1}\textbf{0.9427}&\cellcolor{color1}\textbf{0.8922}&\cellcolor{color1}\textbf{0.9017}&\cellcolor{color1}\textbf{0.9132}&\cellcolor{color1}\underline{0.8721}&\cellcolor{color1}\textbf{0.9249}&\cellcolor{color1}\textbf{0.7653}&\cellcolor{color1}\textbf{0.8433} \\
        &\cellcolor{color2}Auto-Switch (Ours) &\cellcolor{color2}\textcolor{blue!70}{\faCheck}& \cellcolor{color2}\textcolor{red!70}{\faCheck} &\cellcolor{color2}65.5&\cellcolor{color2}\underline{0.5339}&\cellcolor{color2}\textbf{0.9427}&\cellcolor{color2}0.8824&\cellcolor{color2}\textbf{0.9017}&\cellcolor{color2}\textbf{0.9132}&\cellcolor{color2}\textbf{0.8722}&\cellcolor{color2}\underline{0.9193}&\cellcolor{color2}\textbf{0.7653}&\cellcolor{color2}\underline{0.8413}\\
        \bottomrule[1.5pt]
    \end{tabular}}
    \vspace{-3pt} 
    \caption{Main results of merging full-rank task vectors of the RoBERTa models on eight language datasets. The best method is highlighted in bold, and the second-best method is underlined.}
    \label{Roberta}
    \vspace{-15pt}
\end{table*}

\subsection{Merging vision models}

\noindent\textbf{Experimental Settings.}
We follow the setting from \cite{DBLP:journals/corr/abs-2405-17461,DBLP:journals/corr/abs-2406-15479}. For the pre-trained models, we use two variants of CLIP \cite{DBLP:conf/icml/RadfordKHRGASAM21} models' vision encoders: ViT-B/32 and ViT-L/14. We select eight datasets—SUN397 \cite{DBLP:conf/cvpr/XiaoHEOT10}, Cars \cite{DBLP:conf/iccvw/Krause0DF13}, RESISC45 \cite{DBLP:journals/pieee/ChengHL17}, EuroSAT \cite{DBLP:journals/staeors/HelberBDB19}, SVHN \cite{netzer2011reading}, GTSRB \cite{DBLP:conf/ijcnn/StallkampSSI11}, MNIST \cite{DBLP:MNIST}, and DTD \cite{DBLP:conf/cvpr/CimpoiMKMV14}—as our benchmark, with all evaluation metrics being classification accuracy. To ensure fairness, we set the discard ratio to 0.5 for all methods that employ a discard strategy, including ours. We conduct model merging experiments on fully fine-tuned and LoRA fine-tuned low-rank task vectors to thoroughly validate the advantages of our method in both scenarios. Additional training details can be found in the appendix.

\noindent\textbf{Baselines.}
We compare our methods with the following baselines: (1) Pre-trained Models, (2) Individual Models, (3) Traditional MTL
, (4) Weight-Averaging, (5) Task-Arithmetic \cite{DBLP:conf/iclr/IlharcoRWSHF23}, (6) Ties-Merging \cite{DBLP:conf/nips/YadavTCRB23}, (7) DARE Merging \cite{DBLP:conf/icml/Yu0Y0L24}, (8) RegMean \cite{DBLP:conf/iclr/Jin0P023} , (9) Fisher Merging \cite{DBLP:conf/nips/MatenaR22}, (10) AdaMerging \cite{DBLP:conf/iclr/YangW00G0T24}, (11) Twin-merging \cite{DBLP:journals/corr/abs-2406-15479}, (12) EMR-merging \cite{DBLP:journals/corr/abs-2405-17461}.

\noindent\textbf{Main Results.}
Table \ref{ViT-B-32} and Table \ref{LoRA_ViT-B-32} show the results of our experiments of merging full-rank and low-rank task vectors of the ViT-B/32 model on eight tasks. In addition, we list the results of pre-training and fine-tuning as upper and lower bounds on the fusion effects, with the effects of traditional multi-task learning as a comparison. In addition, Table \ref{ViT-B-32} lists the additional storage requirements required for each dynamic merging method (excluding pre-trained weights). 
Observing Table \ref{ViT-B-32} and Table \ref{LoRA_ViT-B-32}, we come up with some interesting conclusions: (1) Due to the fact that dynamic merging methods retain more specific knowledge, they are usually better than static merging methods, but they also create storage burdens, which means there is a trade-off between storage efficiency and performance effectiveness. In the full rank task space, our T-Switch outperforms EMR-merging by 2.24\% and even approaches fine-tuning performance, while in the low-rank task space, it still outperforms EMR-merging by 1.58\%. Our method can still achieve excellent performance even after discarding a large amount of parameter and amplitude information, and its storage is only 12.4\% of EMR. This suggests that the task information within the task vector is highly sparse, meaning that retaining \textbf{less information leads to better model performance}. (2) Our Auto-Switch is a bit lower than the T-Switch in the way. This is because both SUN397 and RESISC45 are scene classification datasets, and their high interclass similarity leads to poor discrimination of KNN on SUN397 and RESISC45, which is the main reason for the decrease in accuracy of the Auto-Switch. As for the other tasks involving target detection or specific object classification tasks, the feature differences are more pronounced, making task types easier to identify and resulting in better classification performance. Overall, our Auto-Switch outperforms Twin-merging by 7.18\% on full-rank task vectors but is 1.03\% lower on low-rank task vectors, while requiring only 1.6\% of Twin-merging's storage.

\subsection{Merging language models}
To more comprehensively verify the generality of T-Switch and Auto-Switch, we merging experiments on models fine-tuned on eight language tasks.

\noindent\textbf{Experimental Settings.}
We follow the setup from \cite{DBLP:journals/corr/abs-2405-17461,DBLP:journals/corr/abs-2406-15479}, using RoBERTa-base \cite{DBLP:journals/corr/abs-1907-11692} as our pre-trained model and fine-tuning it on eight tasks from the GLUE \cite{DBLP:conf/iclr/WangSMHLB19} benchmark: CoLA \cite{DBLP:journals/tacl/WarstadtSB19}, SST-2 \cite{DBLP:conf/emnlp/SocherPWCMNP13}, MRPC \cite{DBLP:conf/acl-iwp/DolanB05}, STS-B \cite{DBLP:journals/corr/abs-1708-00055}, QQP \cite{DBLP:qqp}, MNLI \cite{DBLP:conf/naacl/WilliamsNB18}, QNLI \cite{DBLP:conf/emnlp/RajpurkarZLL16}, and RTE \cite{DBLP:conf/acl/GiampiccoloMDD07}. For evaluation, we apply GLUE’s evaluation metrics: CoLA is evaluated using the Matthews correlation coefficient, STS-B is evaluated using the average of the Pearson and Spearman correlation coefficients, and accuracy is used for the remaining tasks.

\begin{figure}[t] 
    \centering
    \begin{subfigure}{0.48\linewidth}
        \centering
        \includegraphics[width=\linewidth]{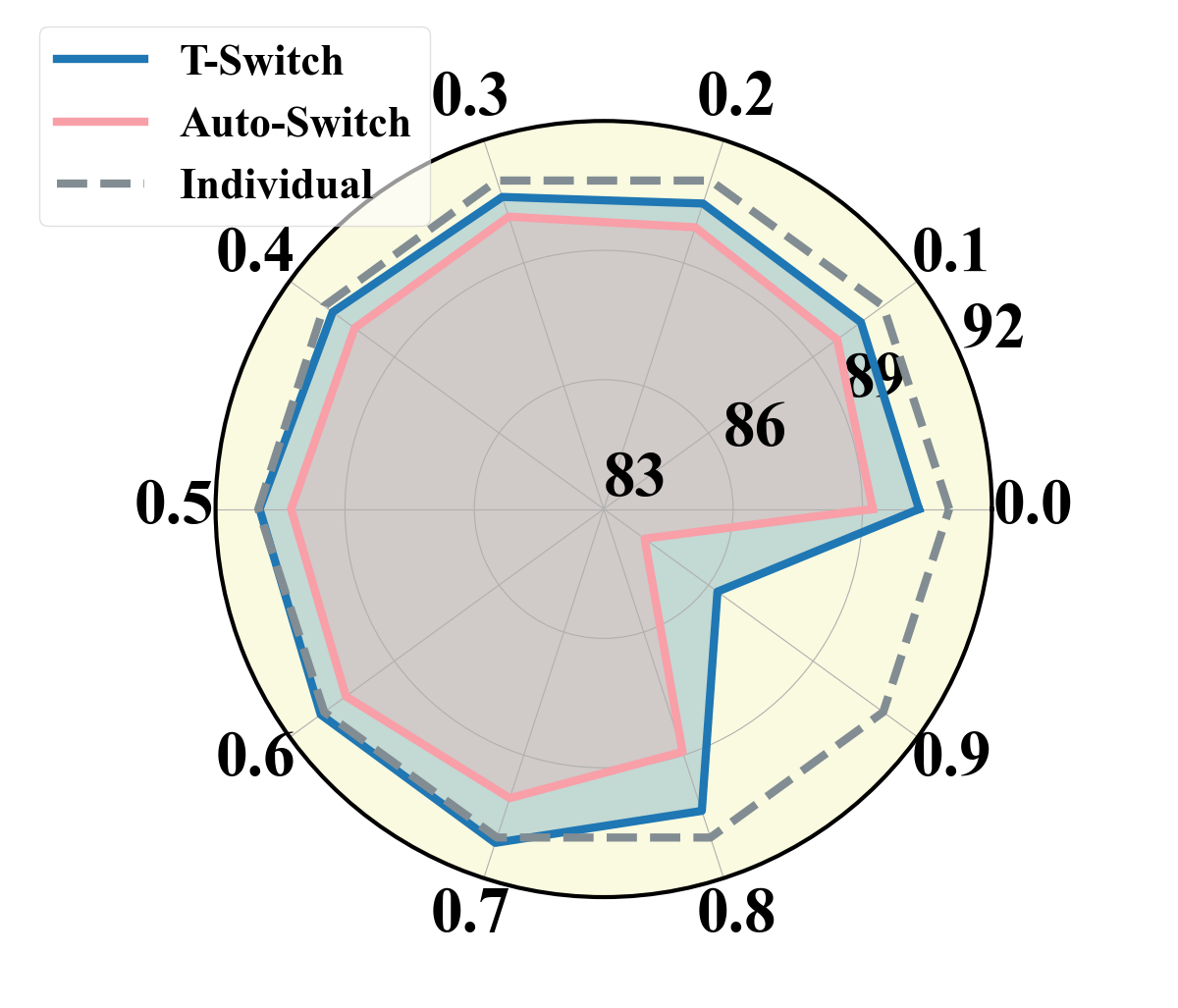}
        \label{fig:sub1}
    \end{subfigure}
    \begin{subfigure}{0.48\linewidth}
        \centering
        \includegraphics[width=\linewidth]{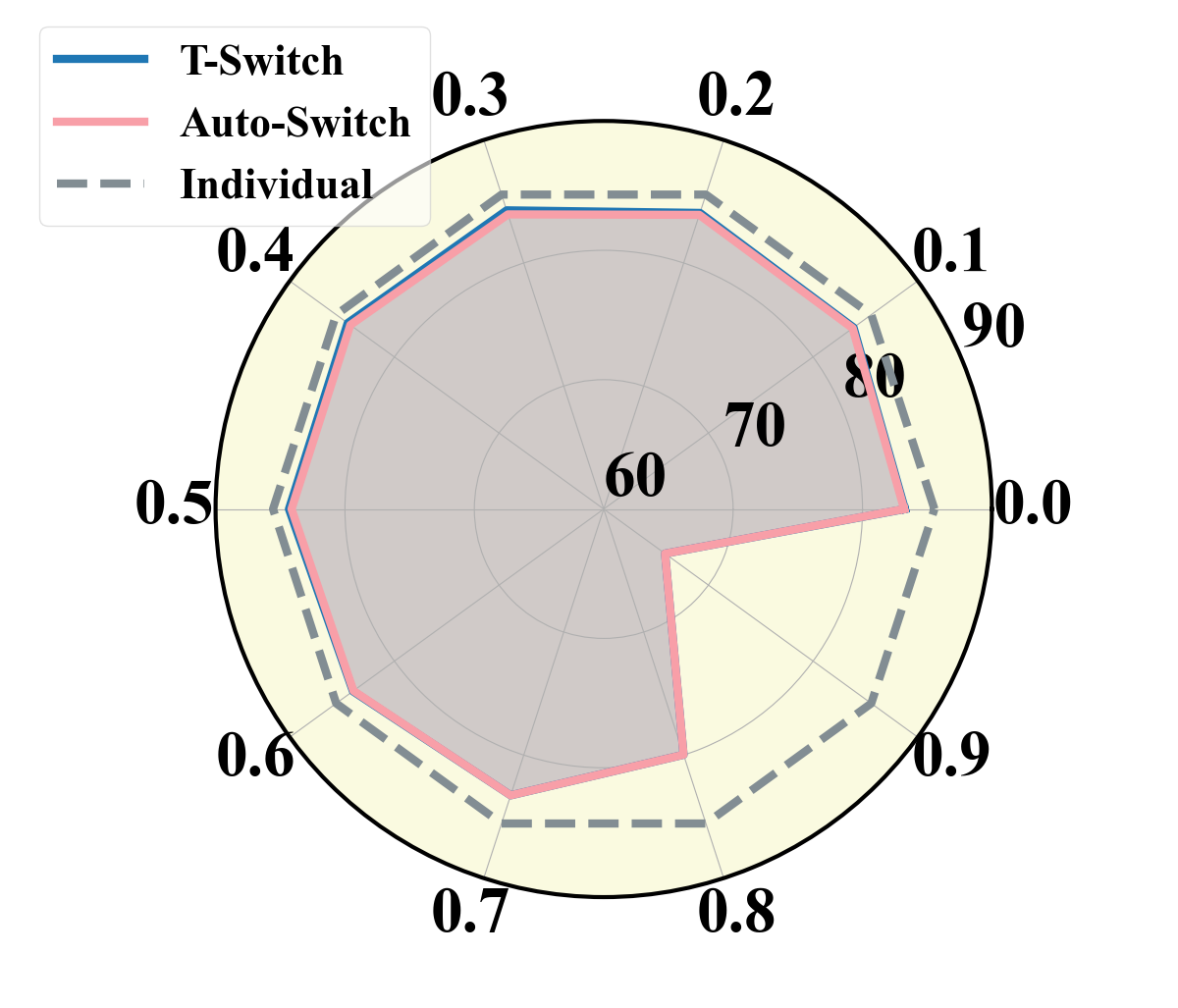}
        \label{fig:sub3}
    \end{subfigure}
    \vspace{-20pt}
    \caption{Merging results ($\%$) with discard ratios ranging from 0.0 to 0.9 across different model. Left: ViT-B/32, Right: RoBERTa.}
    \label{Ablation}
    \vspace{-15pt}
\end{figure}

\noindent\textbf{Baselines.}
We use the same baseline methods as in the merging experiments for vision models. However, since STS-B is a regression task while the others are classification tasks, and Traditional MTL is only applicable to tasks of the same type while AdaMerging is restricted to classification tasks, we excluded these two baselines.

\noindent\textbf{Main Results.}
Table \ref{Roberta} presents the merging performance of full-rank task vectors for all baselines and our method on language tasks, along with the pre-training and fine-tuning performance and the additional storage required by dynamic merging methods. By examining Table \ref{Roberta}, we can draw the following conclusions: (1) Our T-Switch outperforms EMR-merging by 0.0415 points, only 0.0123 points lower than fine-tuning, and our additional storage requirement remains at only 12.5\% of EMR-merging. (2) Given the eight datasets originate from distinct domains with significant differences in data characteristics, our Auto-Switch is only 0.002 points lower than the T-Switch, 0.0311 points higher than Twin-merging, and still requires only 1.7\% of Twin-merging’s storage, demonstrating the generalizability of our method.

\subsection{Ablation Study}
To investigate the impact of the discard ratio on our method, we conduct a series of ablation studies with varying discard ratios. Figure \ref{Ablation} shows the results of merging eight datasets using our method across the ViT-B/32, ViT-L/14 and RoBERTa models, with discard ratio ranging from 0.0 to 0.9. The ablation results on the ViT-L/14 model are provided in the appendix. We discover that, for both vision and language tasks, as the discard ratio increases, our method gradually approaches the fine-tuning performance, only beginning to decline once it surpasses a critical threshold. This aligns well with intuition: as the discard ratio rises, the redundant parameters in the model decrease, reducing interference between different tasks. However, when the discard ratio becomes too high, essential task information starts to be lost, resulting in decreased merging performance. Additionally, we observe an interesting phenomenon: when merging eight vision tasks on the ViT-B/32 model with a discard ratio of 0.7, T-Switch achieves a performance of 91.14\%, surpassing the fine-tuning result of 91.01\% by 0.13\%. On the RoBERTa model, T-Switch attains its highest merging performance of 0.8446 at a discard ratio of 0.4. Given the greater sensitivity of language tasks to parameter adjustments, this result is slightly below the fine-tuning performance of 0.8556.

\section{Conclusion}

In this paper, we observe an impressive conclusion through controlled experiments: discarding parameters with small magnitudes from the task vectors can further improve the performance of both fine-tuned and merged models. Based on this, we propose to binarize the full-precision task vectors, which significantly reduces the storage burden of task parameters while maintaining performance. This leads to the introduction of T-Switch, an efficient dynamic merging method that enables flexible dynamic merging based on binarized task vectors. Furthermore, we introduce Auto-Switch, which automatically combines task switches using a small query set without the need for additional training. Experimental results indicate that both T-Switch and Auto-Switch achieved significant performance improvements on multiple vision and language tasks, requiring only 1-3$\%$ of the storage space compared to full-precision task vectors. Our approach not only enhanced task vector storage efficiency and adaptability but also offered new insights for lightweight storage and deployment of weights in widely used parameter-efficient fine-tuning strategies.
\clearpage
{
    \small
    \bibliographystyle{ieeenat_fullname}
    \bibliography{main}
}

\clearpage
\setcounter{page}{1}
\maketitlesupplementary

\section{Exprimental Details}
To provide a comprehensive overview of the experimental setup, we list the hyperparameter settings for our method and all baseline methods in Table \ref{expriment-settings}.
\begin{table}[h!]
    \centering
    \renewcommand{\arraystretch}{1.0}
    \resizebox{0.45\textwidth}{!}{ 
    \begin{tabular}{l c c c c c c c c}
        \toprule
        Methods& $\alpha$ & $N$ & Scaling Coef & LR & Epochs \\
        \midrule
        Task-Arithmetic& -- & -- & 0.3&--& --  \\
        Ties-Merging&0.5 & -- &  0.3 & --& -- \\
        DARE&0.5 & --& --& --& --\\
        RegMean&-- &256& --& --& -- \\
        Fisher-Merging&--& 4096& -- & --& -- \\
        AdaMerging&-- & --& -- & 1e-3 & 500  \\
        AdaMerging++&0.5 & --& -- & 1e-3 & 500  \\
        Twin-merging&0.5 &  100 & 0.3 &  1e-3 & 10 \\
        EMR-merging&--& -- & -- & --& --  \\
        T-Switch(Ours) &0.5&--& -- & --& -- \\
        Auto-Switch(Ours) &0.5&100& --& -- & -- \\
        \bottomrule[1.5pt]
    \end{tabular}}
    \caption{Hyperparameter settings of our method and all baselines.}
    \label{expriment-settings}
\end{table}

Here, $ \alpha $ represents the discard rate of task vector parameters, $ N $ denotes the number of example samples retained for each task, and "Scaling Coef" refers to the scaling coefficient applied to the merged task vector. LR indicates the learning rate used for training the merging weights or coefficients (e.g., AdaMerging and AdaMerging++) or the task router (e.g., Twin-merging). "Epochs" refers to the additional training epochs required for the merging method. A dash ('-') in the table indicates that the corresponding method does not involve the specified hyperparameter. The hyperparameter settings for all baseline methods follow the configurations provided in the original papers.

\section{Additional Results}

\noindent\textbf{Merging results on the ViT-L/14 model.}
To evaluate the effectiveness of our method in merging larger models, we conducted experiments on eight visual tasks using the ViT-L/14 model. Table \ref{ViT-L-14} shows the combined performance of our method and various baseline methods. Our T-Switch and Auto-Switch achieved the best and second-best results, respectively, significantly outperforming other baseline methods. Notably, T-Switch even surpasses the average performance achieved in the Individual case. This demonstrates that our method retains excellent performance on larger visual models and validates the generalizability of our proposed methods. 
\begin{figure}[htbp] 
    \centering
    \begin{subfigure}{0.46\linewidth}
        \centering
        \includegraphics[width=\linewidth]{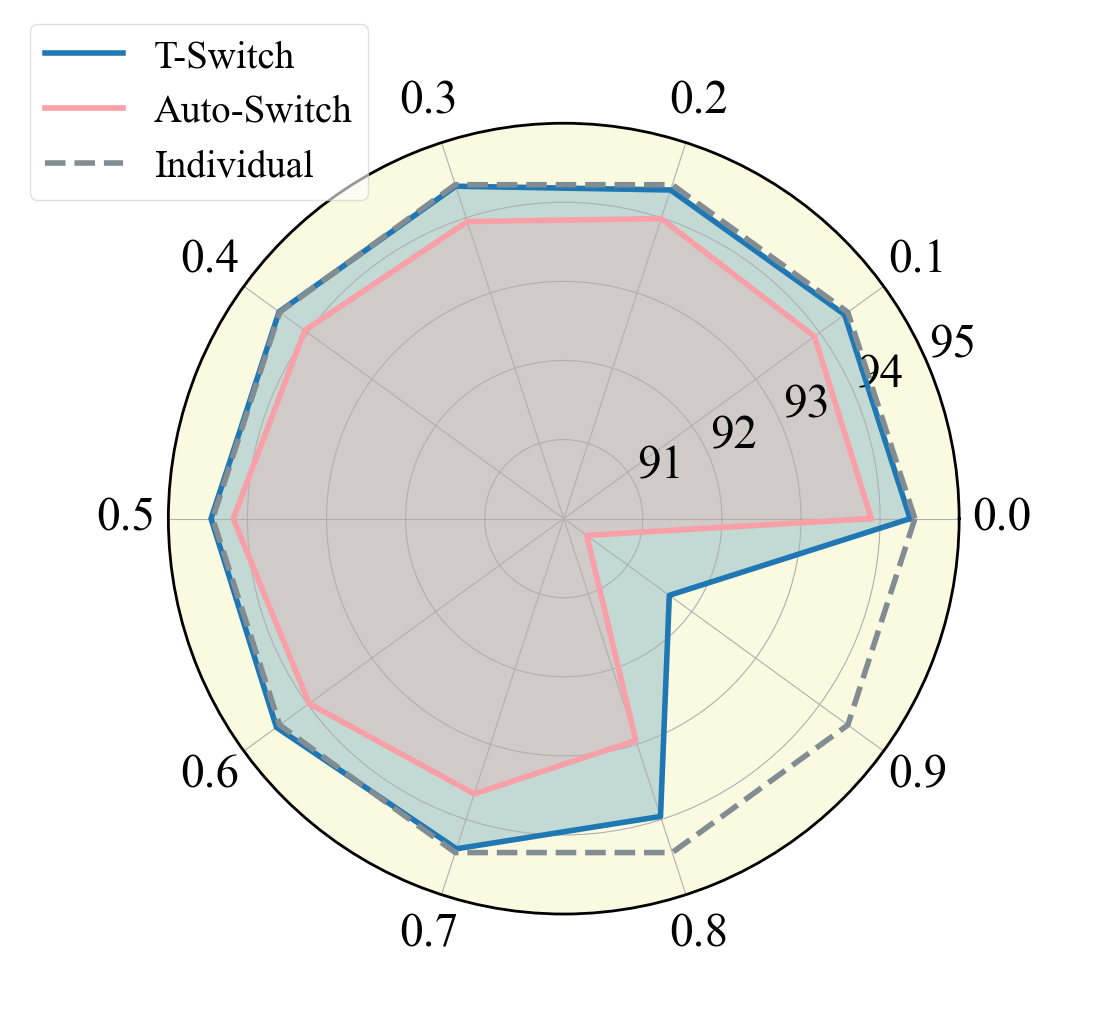}
        \label{fig:sub1}
    \end{subfigure}
    \begin{subfigure}{0.52\linewidth}
        \centering
        \includegraphics[width=\linewidth]{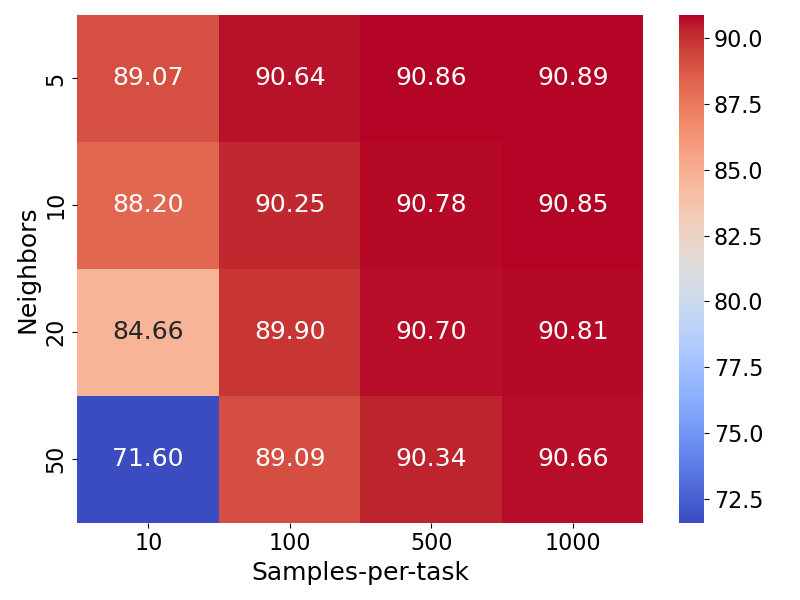}
        \label{fig:sub3}
    \end{subfigure}
    \caption{Additional ablation results. Left: Merging results(\%) with discard ratios ranging from 0.0 to 0.9 on the ViT-L-14 model. Right: Ablation results of the Auto-Switch method merging eight visual tasks on the ViT-B/32 model when the discard ratio is 0.5.}
    \label{Ablation}
\end{figure}

Additionally, we have verified the impact of different dropout rates $\alpha$ on the merging performance of our method on ViT-L by conducting ablation experiments on the ViT-L/14 model with various dropout rates $\alpha$. The left panel of Fig. \ref{Ablation} presents the ablation results for merging eight visual tasks with discard ratio $\alpha$ ranging from 0.0 to 0.9 using the ViT-L/14 model. From the results, we observe a similar phenomenon to that found in the main experiments: as the discard ratio increases, the merging performance of T-Switch improves, even surpassing fine-tuning performance at a discard ratio of 0.6, with noticeable performance degradation only occurring beyond $\alpha=0.7$. As the number of discarded redundant parameters increases, the interference between tasks during merging also decreases. Consequently, the performance of T-Switch and Auto-Switch shows improvement with a moderate increase in $\alpha$.

\begin{table*}[h]
    \centering
    \renewcommand{\arraystretch}{1.0}
    \resizebox{0.95\textwidth}{!}{ \begin{tabular}{c l c c c c c c c c c c c c }
        \toprule
        Type&Methods&Automatic & Example& Storage&SUN397&Cars&RESISC45&EuroSAT&SVHN&GTSRB&MNIST&DTD&AVG \\
        \midrule
        \multirow{3}{*}{--}&Pretrained&--& --& --&66.87&77.94&71.33&62.22&58.45&50.55&76.36&55.37&64.89 \\
        &Individual&--& --& --&84.86&92.39&97.37&99.74&98.11&99.24&99.69&84.15&94.44 \\
        &Traditional MTL&--& --& --&80.80&90.60&96.30&96.30&97.60&99.10&99.60&84.40&93.09 \\
        \hline
        \multirow{8}{*}{\rotatebox{90}{\textbf{Fixed}}}&Weight-Averaging&--&\textcolor{blue!70}{\faTimes}& --&71.10&81.56&82.60&90.63&78.23&70.65&97.01&62.77&79.32 \\
        &Task-Arithmetic&--&\textcolor{blue!70}{\faTimes}& --&73.91&82.13&86.65&92.70&87.91&86.78&98.94&65.64&84.33 \\
        &Ties-Merging&--&\textcolor{blue!70}{\faTimes}& --&73.43&79.75&85.33&91.15&89.98&87.51&99.15&65.21&83.94 \\
        &DARE&--&\textcolor{blue!70}{\faTimes}& --&73.03&82.70&86.19&93.41&85.26&83.48&98.58&65.69&83.54 \\
        &RegMean&--&\textcolor{red!70}{\faCheck}& --&73.04&86.10&88.40&97.52&91.53&89.78&99.0&69.95&86.91 \\
        &Fisher-Merging&--&\textcolor{red!70}{\faCheck}& --&68.11&84.54&75.13&84.11&95.64&91.36&95.56&67.23&82.71 \\
        &AdaMerging&--&\textcolor{red!70}{\faCheck}& --&79.00 &90.30 &90.80 &96.20&93.40 &98.00 &99.00 &79.90 &90.83 \\
        &AdaMerging++&--&\textcolor{red!70}{\faCheck}& --&79.40 &90.30 &91.60&97.40 &93.40 &97.50 &99.00 &79.20 &90.98 \\
        \midrule
        \multirow{4}{*}{\rotatebox{90}{\textbf{Adaptive}}}&Twin-merging&\textcolor{blue!70}{\faCheck}& \textcolor{red!70}{\faCheck}& 10476.1&\underline{84.41}&91.57&\underline{96.95}&\textbf{99.70}&\textbf{98.18}&92.40&99.74&\textbf{84.52}&93.43 \\
        &EMR-merging&\textcolor{red!70}{\faTimes}&\textcolor{blue!70}{\faTimes}&1391.5&83.17 &90.71 &96.78& \textbf{99.70} &97.94 &99.09& 99.69 &82.71 &93.73 \\
        & T-Switch(Ours) &\textcolor{red!70}{\faTimes}&\textcolor{blue!70}{\faTimes} &172& \textbf{84.71} & \textbf{92.54}& \textbf{97.46}&99.67&98.11& \textbf{99.27}& \textbf{99.75}&\underline{84.15}& \textbf{94.46} \\
        & Auto-Switch(Ours) &\textcolor{blue!70}{\faCheck}& \textcolor{red!70}{\faCheck} &174.4&83.27&\underline{92.50}&96.71&99.67&\underline{98.12}&\underline{99.24}&\textbf{99.75}&\underline{84.15}&\underline{94.18}\\
        \bottomrule[1.5pt]
    \end{tabular}}
    \caption{Main results of merging full-rank task vectors of the ViT-L/14 model on eight vision datasets. The best method is highlighted in bold, and the second-best method is underlined.}
    \label{ViT-L-14}
\end{table*}
\noindent\textbf{Ablation of Auto-Switch hyperparameters: samples-per-task $N$ and number of neighbors $C$.}
In our proposed Auto-Switch, there are two additional hyperparameters of this method besides the discard ratio $\alpha$, namely the number of samples retained for each task $N$ and the number of neighbors $C$. To verify the impact of these two hyperparameters on Auto-Switch, we conduct ablation experiments on the ViT-B/32 model. The results shown in Fig. \ref{Ablation} indicates that: 1) As the number of neighbors $C$ increases, the model's merging performance tends to decrease. This happens because, when selecting neighbors from the local vicinity of the input sample, an increase in the number of neighbors (especially when it approaches the total number of samples) can introduce many distant, irrelevant samples. These distant samples can negatively impact classification accuracy, thereby reducing the effectiveness of the merging process. Therefore, selecting an appropriate number of neighbors $C$, is crucial. 2) As the number of samples per task $N$ increases, the model's merging performance improves significantly. This is because more samples help to concentrate the features of each dataset, which in turn enhances the stability of neighbor selection.

\end{document}